\documentclass[10pt,twocolumn,letterpaper]{article}
\usepackage{iccv}
\usepackage{times}
\iccvfinalcopy
\usepackage{epsfig}
\usepackage{graphicx}
\usepackage{amsmath}
\usepackage{amssymb}
\usepackage{booktabs}
\usepackage{wrapfig}
\usepackage{multirow}
\usepackage{bm}
\usepackage{enumitem}
\usepackage{pifont}% http://ctan.org/pkg/pifont
\usepackage{amsmath}
\usepackage{amssymb}
\usepackage{booktabs}
\usepackage{enumitem}
\usepackage{multirow}
\usepackage{makecell}
\usepackage{graphicx}
\usepackage{diagbox}
\usepackage{tabu}
\usepackage{rotating}
\usepackage{array,multirow}
\usepackage{float}
\usepackage{enumitem}
\usepackage{wrapfig}
\usepackage{pifont}
\usepackage{algorithm}
\usepackage{color}
\usepackage{algpseudocode} 
\usepackage[table]{xcolor}
\usepackage{booktabs}
\usepackage{tabularx}

\newcommand{\parsection}[1]{\vspace{4pt}\noindent\textbf{#1:}}

\newcommand{\cmark}{\ding{51}}%
\newcommand{\xmark}{\ding{55}}%
\usepackage{diagbox}

\newcommand{\inlineimg}[1]{\raisebox{-0.1\baselineskip}{\includegraphics[height=0.99\baselineskip]{#1.png}}}

\usepackage[pagebackref=true,breaklinks=true,letterpaper=true,colorlinks,bookmarks=false]{hyperref}

\usepackage[capitalize]{cleveref}
\crefname{section}{Sec.}{Secs.}
\Crefname{section}{Section}{Sections}
\Crefname{table}{Table}{Tables}
\crefname{table}{Tab.}{Tabs.}

\def\iccvPaperID{5891} % *** Enter the ICCV Paper ID here

\ificcvfinal\fi

\begin{document}

%%%%%%%%% TITLE
\title{Invariant Training 2D-3D Joint Hard Samples for Few-Shot Point Cloud Recognition}

\author{
\textbf{Xuanyu Yi}\textsuperscript{1}, \textbf{Jiajun Deng}\textsuperscript{2}, \textbf{Qianru Sun}\textsuperscript{5}, \textbf{Xian-Sheng Hua}\textsuperscript{3},\\
\textbf{Joo-Hwee Lim}\textsuperscript{4}, \textbf{Hanwang Zhang}\textsuperscript{1}\\
{\small \textsuperscript{1}Nanyang Technological University, \textsuperscript{2}The University of Sydney} \\
{\small \textsuperscript{3}Terminus Group, \textsuperscript{4}Institute for Infocomm Research,}
{\small \textsuperscript{5}Singapore Management University}\\
{\tt\small xuanyu001@e.ntu.edu.sg, jiajun.deng@sydney.edu.au, qianrusun@smu.edu.sg }\\
{\tt\small xshua@outlook.com, joohwee@i2r.a-star.edu.sg, hanwangzhang@ntu.edu.sg}
}

\maketitle
% Remove page # from the first page of camera-ready.
\ificcvfinal\thispagestyle{empty}\fi

%%%%%%%%% ABSTRACT
\begin{abstract}
  We tackle the data scarcity challenge in few-shot point cloud recognition of 3D objects by using a joint prediction from a conventional 3D model and a well-trained 2D model. Surprisingly, such an ensemble, though seems trivial, has hardly been shown effective in recent 2D-3D models. We find out the crux is the less effective training for the ``joint hard samples'', which have high confidence prediction on different wrong labels, implying that the 2D and 3D models do not collaborate well. To this end, our proposed invariant training strategy, called \textsc{InvJoint}, does not only emphasize the training more on the hard samples, but also seeks the invariance between the conflicting 2D and 3D ambiguous predictions. \textsc{InvJoint} can learn more collaborative 2D and 3D representations for better ensemble.  Extensive experiments on 3D shape classification with widely adopted ModelNet10/40, ScanObjectNN and Toys4K, and shape retrieval with  ShapeNet-Core validate the superiority of our \textsc{InvJoint}. Codes will be publicly Available~\footnote{\url{https://github.com/yxymessi/InvJoint}}.
\end{abstract}

%%%%%%%%% BODY TEXT

\section{Introduction}
\label{sec:intro}

\begin{figure}[t]
   \begin{minipage}[b]{1.0\linewidth}
   \centerline{\includegraphics[width =9.2cm]{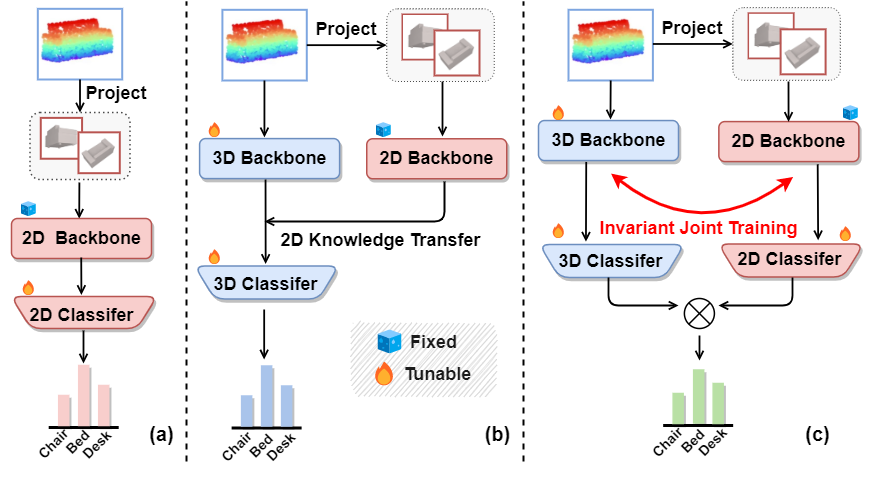}}
   \end{minipage}
   \caption{Comparisons of our framework with existing 2D-3D methods, which can be categorized into (a) Directly projecting point cloud into multi-view images as inputs, and then fine-tuning the 2D models with a frozen backbone. (b) Indirectly leveraging the 2D pretrained knowledge as a constraint or supervision, transferring them via knowledge distillation or contrastive learning, and then only using the optimized 3D pathway for prediction. (c) In contrast, our \textsc{InvJoint}, based on ensemble paradigm, makes the best of the 2D and 3D worlds by joint prediction in inference.}
   \label{fig:1}
\end{figure}

As the point cloud representation of a 3D object is sparse, irregularly distributed, and unstructured, a deep recognition model requires much more training data than the 2D counterpart~\cite{he2016deep,huang2019convolutional}. Not surprisingly, this makes few-shot learning even more challenging, such as recognizing a few newly-collected objects in AR/VR display~\cite{jang2019progress,sun2023trosd} and  robotic navigation~\cite{adams2012robotic}. Thanks to the recent progress of large-scale pre-trained multi-modal foundation models~\cite{radford2021learning, li2022blip, lu2019vilbert}, the field of 2D few-shot or zero-shot recognition has experienced significant improvements. Therefore, as shown in Figure~\ref{fig:1}(a), a straightforward solution for 3D few-shot recognition is to project a point cloud into a set of multi-view 2D images~\cite{goyal2021revisiting}, through rendering and polishing~\cite{wang2022p2p}, and then directly fed the images into a well-trained 2D model~\cite{zhang2022pointclip}.

\begin{figure*}[t]
  \begin{minipage}[b]{1.0\linewidth}
  \centerline{\includegraphics[scale = 0.58]{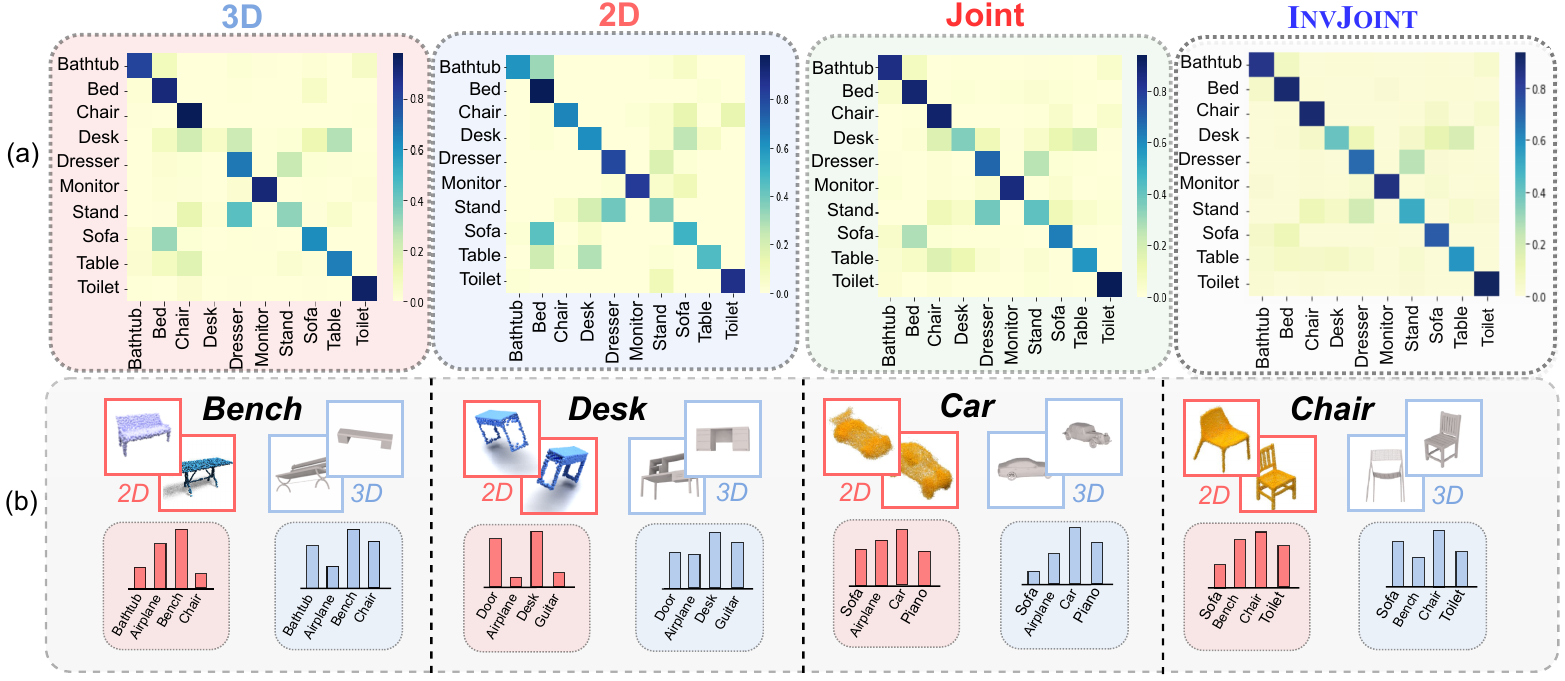}}
  \end{minipage}
  \caption { (a) 3D and 2D models are confused by different classes, thus a simple late fusion cannot turn the joint confusion matrix more diagonal. (b) Qualitative examples of \textbf{joint hard samples} with their logits distribution.
 }
  \label{fig:2}
\end{figure*}

Although effective, the projected images are inevitably subject to incomplete geometric information and rendering artifacts. To this end, as shown in Figure~\ref{fig:1}(b), another popular solution attempts to take the advantage of both 2D and 3D by transferring the 2D backbone to the 3D counterpart via knowledge distillation~\cite{yan2022let}, and then use the 3D pathway for  final recognition. So far, you may ask: as the data in few-shot learning is already scarce, during inference time, why do we still have to choose one domain 
or the other? Isn't it common sense to combine them for better prediction? In fact, perhaps for the same reason, the community avoids answering the question---our experiments (Section~\ref{sec:4}) show that a naive ensemble, no matter with early or late fusion, is far from being effective as it only brings marginal improvement.

To find out the crux, let's think about in what cases, the ensemble can correct the individually wrong predictions by joint prediction, \eg, if the ground-truth is ``Bench'' and neither 2D nor 3D considers ``Bench'' as the highest confidence, however, their joint prediction peaks at ``Bench''. The cases are: 1) the ground-truth confidence of the two models cannot be too low, and 2) that of  the other classes cannot be too high. In one word, 2D and 3D are collaborative. However, as shown by the class confusion matrices of training samples in Figure~\ref{fig:2}(a), since 2D and 3D are confused by different classes, their ensemble can never turn the matrix into a more diagonal one. This implies that their joint prediction in inference may be still wrong. 

Therefore, the key is to make the joint confusion matrix more diagonal than each one. To this end, we focus on the \textbf{joint hard samples}, which have high confidence prediction on different wrong labels respectively. See Figure~\ref{fig:2}(b) for some qualitative examples, exhibiting a stark difference in logits distribution among modalities. However, simply re-training them like the conventional hard negative mining~\cite{suh2019stochastic, shrivastava2016training} is less effective because the joint training is easily biased towards the ``shortcut'' hard samples in one domain. For example, if the 3D model has a larger training loss than 2D, probably due to a larger sample variance~\cite{zhou2021machine}, which is particularly often in few-shot learning, the joint training will only take care of 3D, leaving  2D still or even more confused. In Section~\ref{sec:5}, we provide a perspective on joint hard samples from the view of probability theory, while the Venn Diagram perspective in \textit{Appendix}.
% Due to limited space, the probability theory as well as the Venn Diagram perspective of \textbf{joint hard samples} are given in \textit{Appendix Part B}.

\begin{figure}[t]
   \begin{minipage}[b]{1.0\linewidth}
   \centerline{\includegraphics[scale = 0.39]{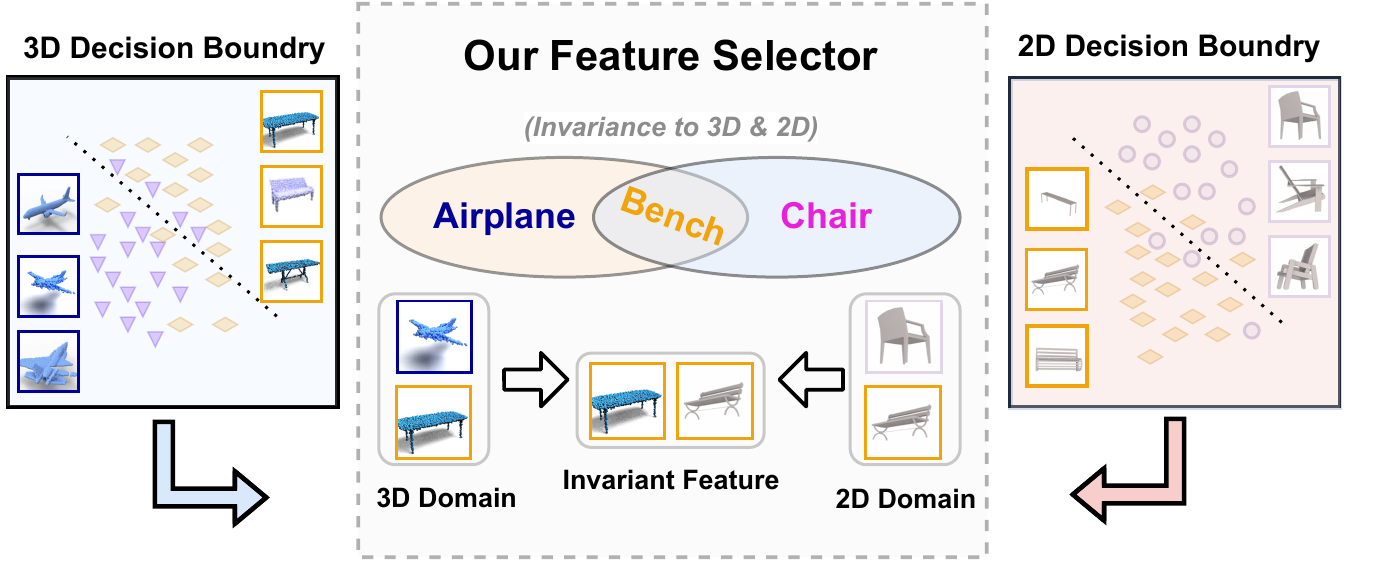}}
   \end{minipage}
   \caption {Illustration of the invariant training idea. Given the predictions of a ``Bench'' sample in both domains, the invariance selector removes the conflict confusion ( ``Chair'' and ``Airplane'') and keeps the common ``Bench''. }
   \label{fig:3}
\end{figure}

By consolidating the idea of making use of joint hard examples for improving few-shot point cloud recognition, we propose an \textbf{invariant training} strategy. As illustrated in Figure~\ref{fig:3}, if a sample ground-truth is ``Bench'' and 2D prediction is confused between ``Bench'' and ``Chair'', while the 3D counterpart is uncertain about ``Bench'' and ``Airplane'', the pursuit of invariance will remove the variant ``Chair'' and ``Airplane', and eventually keep the common ``Bench'' in each model. Specifically, we implement the strategy as \textsc{InvJoint}, which has two steps to learn more collaborative 2D and 3D representations (Section~\ref{sec:3.3}).
\textbf{Step 1}:  it selects those joint hard samples by firstly fitting a Gaussian Mixture Model of sample-wise loss, and then picking them according to the fused logit distribution. \textbf{Step 2}: A joint learning module focusing on the selected joint hard samples effectively capture the collaborative representation across domains through an invariant feature selector. After the \textsc{InvJoint} training strategy, a simple late-fusion technique can be directly deployed for joint prediction in inference (Section~\ref{sec:3.4}). Figure~\ref{fig:2}(a) shows that the joint confusion matrix of training data is significantly improved after \textsc{InvJoint}.
% Concretely, rich color and fine-grained texture are easily obtained in 2D images, but they are ambiguous in depth and shape sensing

We conduct extensive few-shot experiments on several synthetic~\cite{wu20153d,stojanov2021using,chang2015shapenet} and real-world~\cite{uy2019revisiting} point cloud 3D classification datasets. \textsc{InvJoint} gains substantial improvements over existing SOTAs. Specifically, on ModelNet40, it achieves an absolute improvements of \textit{6.02}\% on average and \textit{15.89}\% on 1-shot setting compared with PointCLIP~\cite{zhang2022pointclip}. In addition, the ablation studies demonstrate the component-wise contributions of \textsc{InvJoint}.
In summary, we make three-fold  contributions:
\begin{itemize}
\setlength\itemsep{0em}
\item We propose \textsc{InvJoint} that aims to make the best of the 2D and 3D worlds. To the best of our knowledge, it is the first work that makes 2D-3D ensemble work in point cloud 3D few-shot recognition.

\item We attribute the ineffective 2D-3D ensemble  to the \textit{``joint hard samples''}. \textsc{InvJoint} exploits their 2D-3D conflicts to remove the ambiguous predictions.

\item \textsc{InvJoint} is a plug-and-play training module whose potential could be further unleashed with the evolving backbone networks.

\end{itemize}

\section{Related Work}
Point cloud is the prevailing representation of 3D world. The community has proposed various deep neural networks for point clouds, including convolution-based~\cite{xu2018spidercnn,komarichev2019cnn,li2018pointcnn,deng2021voxel}, graph-based~\cite{wang2019dynamic,zhou2021adaptive,li2021towards},  MLP-based~\cite{qi2017pointnet,qi2017pointnet++,ma2022rethinking,qian2022pointnext}, and the recently introduced Transformer-based~\cite{zhao2021point,yu2022point,guo2021pct}. Despite the fast progress, the performance of these models is limited due to the lack of a properly pre-trained backbone for effective feature representation.  To this end, three main directions are explored: 1) intra-modality unsupervised representation learning, 2) project-and-play by 2D networks, 3) 2D-to-3D knowledge transfer.
% \dnote{These three names should be further discussed.}

\parsection{Point Cloud Unsupervised Feature Learning}
The early work PointContrast~\cite{xie2020pointcontrast} establishes the correspondence between points from different camera views and performs point-to-point contrastive learning in the pre-training stage. Besides contrastive learning, data reconstruction is also explored. OcCo~\cite{wang2021unsupervised} learns point cloud representation by developing an autoencoder to reconstruct the scene from occluded inputs.
However, they generalize poorly to downstream tasks due to the relatively small pre-training datasets.

\parsection{Project-and-Play by 2D Networks} 
The most straightforward way to make use of 2D networks for 3D point cloud understanding is to transfer point clouds into 2D images. Pioneered by MVCNN~\cite{su2015multi} that uses multi-view images rendered from pre-defined camera poses and produces global shape signature by performing cross-view max-pooling, the follow-up works are mainly devoted to more sophisticated view aggregation techniques~\cite{hamdi2021mvtn}. However, the 2D projection inevitably loses 3D structure and thus leads to sub-optimal 3D recognition. 

\parsection{2D-to-3D Knowledge Transfer} It transfers knowledge from a well-pretrained 2D image network to improve the quality of 3D representation via cross-modality learning.  Given point clouds and images captured in the same scene, PPKT~\cite{liu2021learning} first projects 3D points into images to establish the correspondence, and then performs cross-modality contrastive learning in a pixel-to-point manner. CrossPoint~\cite{afham2022crosspoint} proposes a self-supervised joint learning framework that boosts feature learning of point clouds by enforcing both intra- and inter-modality correspondences. The most related work to ours is PointCLIP~\cite{zhang2022pointclip}, which exploits an off-the-shelf image visual encoder pretrained by CLIP~\cite{radford2021learning} to address the problem of few-shot point cloud classification. Different from PointCLIP which directly fine-tunes 2D models for inference, our proposed \textsc{InvJoint} significantly improves 2D-3D joint prediction by invariant training on joint hard samples.

\begin{figure*}[t]
  \begin{minipage}[b]{1.0\linewidth}
  \centerline{\includegraphics[width=170mm]{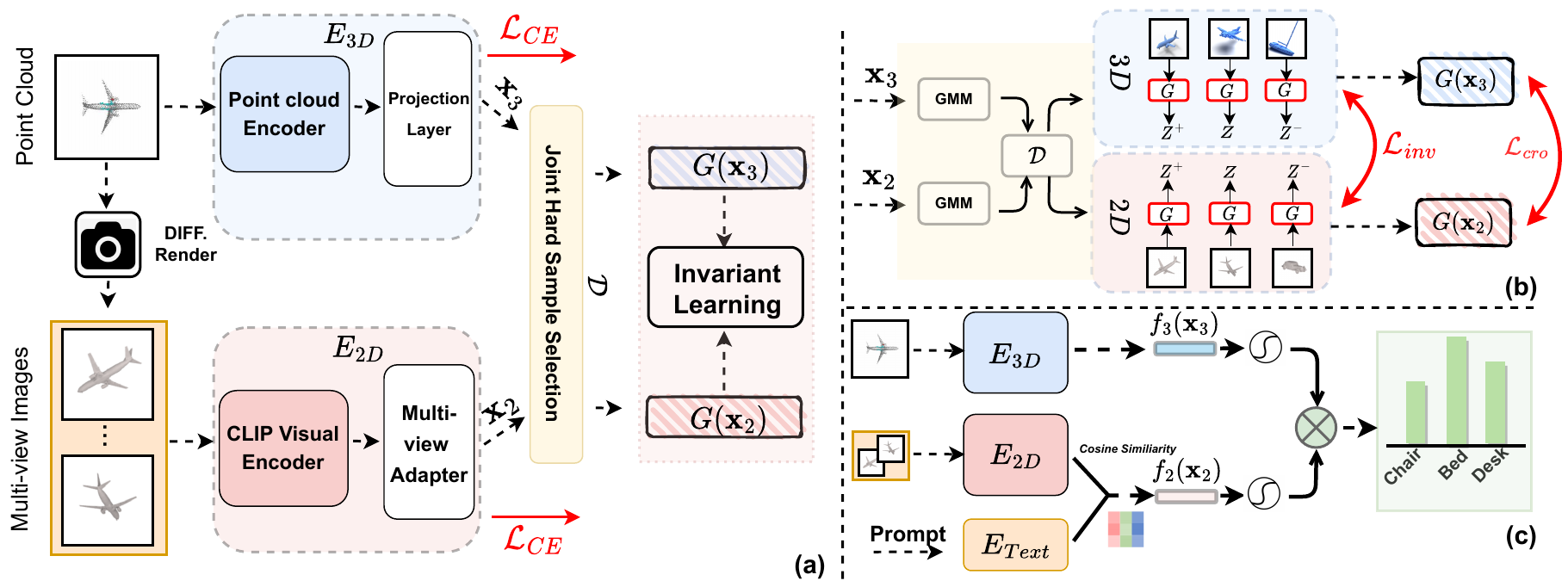}}
  \end{minipage}
  \caption{(a) The training pipeline of \textsc{InvJoint}. $E_{\text{3D}}$, $E_{\text{2D}}$ (including the renderer), and $G$ are trainable parameters. (b) The zoom-in diagram of Invariant Learning. Note that $\mathcal{L}_{\text{inv}}$ only trains $G$. (c) The inference pipeline, where \inlineimg{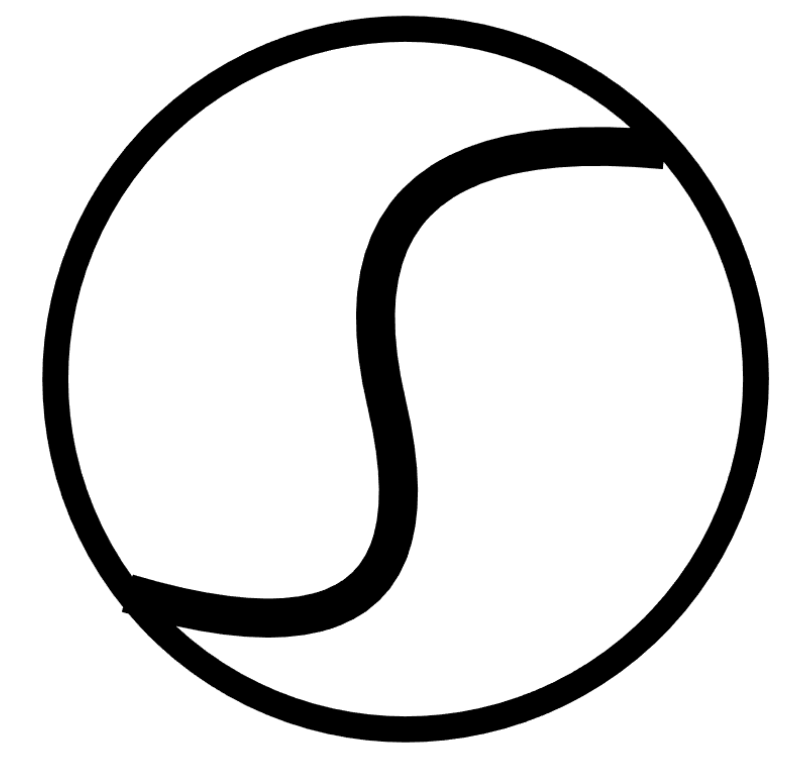} denotes the $\text{Softmax}$ layer.}
  \label{fig:4}
\end{figure*}

\section{\textsc{InvJoint}}
\textsc{InvJoint} is an invariant training strategy that selects and then trains 2D-3D joint hard samples for few-shot point cloud recognition by using 2D-3D joint prediction. The overview of \textsc{InvJoint} is illustrated in Figure~\ref{fig:4}. 
Given 3D point clouds, we first perform image rendering to produce a set of 3D-projected multi-view images as the corresponding 2D input. Then, the point clouds and multi-view images are respectively fed into the 3D and 2D branches for modality-specific feature encoding (Section~\ref{sec:3.2}).
Next, we select joint hard samples  and feed them into the invariant learning module for better collaborative 2D-3D features (Section~\ref{sec:3.3}). At the inference stage, we introduce a simple fusion strategy for joint prediction (Section~\ref{sec:3.4}).

\subsection{Multi-modality Feature Encoding}
\label{sec:3.2}

\parsection{Point Cloud Feature Encoder} In our 3D branch, we extract the geometric features  from point clouds input with the widely adopted DGCNN~\cite{wang2019dynamic}. 
% [INTRODUCE HOW YOU GET THIS? PRE-TRAINED ON WHAT, FREEZE OR NOT?]
Then a trainable projection layer is applied for feature dimension alignment with the 2D feature introduced later. We denote the encoder and its trainable parameters as $E_{\text{3D}}$, and its output feature as $\mathbf{x}_{3}$.

\parsection{Image Feature Encoder} Since 3D-2D input pairs are not always available, to improve the applicability of our method, we adopt the differentiable rendering technique to generate photo-realistic images for the 2D views. Specifically, an alpha compositing renderer~\cite{wiles2020synsin} is deployed with trainable parameters of cameras for optimized recognition. After obtaining the rendered multi-view images, we feed them into the frozen CLIP~\cite{radford2021learning} visual encoder (\ie, the pre-trained ViT-B model) with an additional trainable linear adapter~\cite{gao2021clip} to narrow the gap between the rendered images and the original CLIP images. We denote the image encoder as $E_{\text{2D}}$ and its output feature as $\mathbf{x}_{2}$.

\subsection{Invariant Joint Training} 
\label{sec:3.3}
As we discussed in Section~\ref{sec:intro}, due to the feature gap between  $\mathbf{x}_{2}$ and  $\mathbf{x}_{3}$, there are joint hard samples preventing the model from learning collaborative 2D-3D features.  As illustrated in Figure~\ref{fig:4}(b), our invariant joint training contains the following two steps, performing in an \textit{iterative} pattern:

\noindent\textbf{Step 1: Joint Hard Sample Selection}. We first conduct Hard Example Mining (\textbf{HEM}) in each modality, then combining them subject to a joint prediction threshold. 
\vspace{-0.03cm}

In a certain modality, one is considered as hard sample if its training loss is larger~\cite{hermans2017defense,li2020dividemix} than a pre-defined threshold since deep networks tend to learn from easy to hard~\cite{shrivastava2016training}. In particular, we adopt a two-component Gaussian Mixture Model (GMM) ~\cite{xuan2001algorithms} $P(g  \! \mid \! \mathcal{L}_{CE})$ to normalize the per-sample cross-entropy (CE) loss distribution for each modality respectively: $\mathcal{D}_{i\in\{2, 3\}} = \{\mathbf{x}_i  \mid P\left( g \mid \mathcal{L}_{CE}(\mathbf{x}_i)\right) < p_i\}$
% \begin{equation}
% \label{eq.1}
% \mathcal{D}_{i\in\{2, 3\}} = \{\mathbf{x}_i  \mid P\left( g | \mathcal{L}_{CE}(\mathbf{x}_i)\right) < p_i\},
% \end{equation}
, where g is the Gaussian component with smaller mean, $p_{i}$ is a probability threshold. 
Subsequently, the joint hard samples are chosen based on two criteria:~\footnote{
In order to iteratively capture the joint hard samples and avoid over-fitting the training set: The parameter threshold $r_{1}$, $r_{2}$ is dynamically determined by an overall controlled ratio of observed distribution.}
% Different occlusion-based~\cite{xx} augmentations are conducted to gradually increase the complexity of training samples. 
\vspace{-2mm}
\begin{itemize}
\setlength\itemsep{0em}

\item (1) \textit{The joint prediction has high confidence on wrong labels}, \ie, the sum of 2D and 3D logits in most likely non-ground-truth class is large : $\mathcal{D} = \{(\mathbf{x}_{2}, \mathbf{x}_{3}\in\mathcal{D}_{{2}\cup {3}}) \mid \max_{i\neq gt} f^i_{2}(\mathbf{x}_{2})+f^i_{3}(\mathbf{x}_{3}) > r_{1}\}$, where $f^i(\bm{x})$ denotes the logits output of the $i$-th class.

\vspace{-2mm}

\item (2) \textit{The discrepancy between 2D and 3D logits is apparent}, \ie, the top-5 categories ranks from 2D and 3D logits show inconsistency:
$\mathcal{D} = \{(\mathbf{x}_{2}, \mathbf{x}_{3}\in\mathcal{D}_{{2}\cup {3}}) \mid  ||\text{top}k(f_{2}(\mathbf{x}_{2}), 5)) \cap \text{top}k(f_{3}(\mathbf{x}_{3}), 5)|| < r_{2} \} $, 
where $\text{top}k(f(\mathbf{x}), 5))$ signifies the set of top-5 category indices based on output logit confidence.

\end{itemize}

\vspace{-2mm}

\noindent\textbf{Step 2: Cross-modal Invariance Learning}. The goal of this step is to acquire a non-conflicting feature space by reconciling the 2D-3D features of samples in $\mathcal{D}$. Meanwhile, we also don't want the purist of invariance---seeking common features---to negatively affect the complementary nature of the 2D-3D representation. Therefore, we devise a gate function $G$ which works as a soft mask that applies element-wise multiplication to select the non-conflicting features for each modality, \eg, $G(\mathbf{x}_{2})$ for 2D and $G(\mathbf{x}_{3})$ for 3D, and the following invariance training only tunes $G$ while freezing the feature extractors $E_{\text{3D}}$ and $E_{\text{2D}}$.

Inspired by invariant risk minimization (IRM)~\cite{arjovsky2019invariant}, 
we consider the point cloud feature $\bm{x}_{3}$ and the image features $\bm{x}_{2}$ as two environments, and propose the modality-wise IRM. Due to IRM essentially regularizes the model to be equally optimal across environments, \ie, modalities, we can learn a unified gate function $G$ to select non-conflicting features for each modality by:
% Then we learn the gate function $G$ to select non-conflicting features, thus alleviating the ambiguity for ``joint hard samples''.
% IRM essentially regularizes $G$ to be equally optimal across environments: 
\begin{equation}
    \label{eq.3}
    \begin{split}
         \min_{G} \;  & \sum_{\mathbf{x_{e}}\in \{\mathbf{x}_{2},\mathbf{x}_{3}\}} R_e(\mathbf{x}_{e},y;G) \\
        \text{s.t.} \; & G \in \arg\min_{G} R_e(\mathbf{x}_{e},y;G), \forall \mathbf{x}_{e} \in \{\mathbf{x}_{2},\mathbf{x}_{3}\},
    \end{split}
\end{equation}
where $R_e(\mathbf{x}_{e},y;G)$ is the empirical risk under $e$,  $G$ denotes a learnable mask layer multiplied on $\mathbf{x}_{e}$.

% $\phi_{e}(x)$ denotes the 2D / 3D encoded features; $ R^e(x,y;g(\phi_{e}(x)))$ is the risk under environment $e$; $g \in \arg\min_{g} R^e(x,y;g(\phi_{e}(x)))\text{ for all } e\in \mathcal{E}$ means that the invariant feature selector $g$ should minimize the risk under all environments simultaneously. The detailed process of invariant environmental learning is as below:

% \begin{equation}
%     \label{eq.2}
%     \begin{split}
%          \min_{G} \;  & \sum_{e\in \mathbb{E}} R_e(x,y;G(\mathcal{F}_e)) \\
%         \text{s.t.} \; & G \in \arg\min_{G} R_e(x,y;G(\mathcal{F}_e)), \forall e\in \mathbb{E},
%     \end{split}
% \end{equation}
% where $R_e(x,y;G(\mathcal{F}_e))$ is the empirical risk under $e$. 

% \dnote{To mention the benefit of using this regularizer in one sentence.} 

% \dnote{To mention the aforementioned regularizer or the invariant joint training is based on joint hard samples.}  

% \dnote{To tell the relation between using the feature selector (select modality-invariant class features) and alleviate attribute bias}

% \begin{equation}
%  \label{eq:2}
%  \scriptsize
% \mathcal{L}_{inv} = \sum_{e \in \{\small{\text{3D},\text{2D}\}}} \mathcal{L}_{e}\left(G(\mathcal{F}_{e}) \cdot \theta, y\right)+\lambda\left\|\nabla_{\theta=1} \mathcal{L}_{e}\left(G(\mathcal{F}_{e}) \cdot \theta, y\right)\right\|_{2}^{2}
%  \end{equation}
 Particularly, we implement $R_e(\cdot)$ as our modality-wise IRM by \textit{supervised} InfoNCE loss~\cite{khosla2020supervised}:
%  our modality-wise IRM  is implemented on the contrastive objective, which can be formulated as:
\begin{equation}
\small
    \label{eq.4}
\mathcal{L}_{e}(\boldsymbol{z}_{e}, \theta) = -\log \frac{\exp \left(\boldsymbol{z}_{e}^{\mathrm{T}} \boldsymbol{z}_{e}^{+} \cdot \theta\right)}{\exp \left(\boldsymbol{z}_{e}^{\mathrm{T}} \boldsymbol{z}_{e}^{+} \cdot \theta\right)+\sum_{\boldsymbol{z}_{e}^{-} } \exp \left(\boldsymbol{z}_{e}^{\mathrm{T}} \boldsymbol{z}_{e}^{-} \cdot \theta\right)},
\end{equation}

where $z_{e} = G(\bm{x}_{e})$, which is a element-wise product. To ensure the sufficient labeled samples for positive discrimination, we follow the common practice to utilize regular spatial transformations, \emph{e.g.}, rotation, scaling and jittering  as the augmented point cloud in $\mathbf{x}_{3}$; we consider the different rendering views as the augmented images in $\mathbf{x}_{2}$. Therefore, the augmented $\mathbf{x}_e$ in the same class are taken as positive $\boldsymbol{z}_{e}^{+}$, while the representation of other classes as negative $\boldsymbol{z}_{e}^{-}$ in both modalities respectively. In this way, $G$ is optimized through the proposed modality-wise IRM loss:
\begin{equation}
 \label{eq:5}
 \small
\mathcal{L}_{\text{inv}} = \! \! \sum_{\mathbf{x}_{e} \in \{\small{\mathbf{x}_{2}, \mathbf{x}_{3}\}}} \mathcal{L}_{e}\left(G(\mathbf{x}_{e}) , \theta \right)+\lambda\left\|\nabla_{\theta=1} \mathcal{L}_{e}\left(G(\mathbf{x}_{e}) , \theta\right)\right\|_{2}^{2},
 \end{equation}

where $\lambda$ is a trade-off hyper-parameter; $\theta$ is a dummy classifier to calculate the gradient penalty across modality, which encourages $G$ to select the non-conflicting features.
% As a result, the influence of modality bias is eliminated, leading to the acquisition of a non-conflicting feature space $G(\mathbf{x}_{e})$ for further cross-modality alignment.
% Though different losses may be induced between modalities, $\mathcal{W(\cdot)}$ must regularize them optimal simultaneously in the same way by using the shared dummy classifier $\theta$.
% By this way, the feature selector $\mathcal{W(\cdot)}$ is optimized through the proposed modality-wise IRM loss $\mathcal{L}_{inv}$ from Eq.~\eqref{eq:2}. 

% In practice

% In the modality-invariant feature space, the objective is to maximize the cosine similarity of $\mathbf{z}_{i}^{e_{1}}$ and $\mathbf{z}_{i}^{e_{2}}$, which denote the 3D/2D representation for the same sample

\subsection{Overall Loss}

% \noindent\textbf{Modality-invariant Feature Encoders.} 

% \noindent\textbf{Cross-entropy Loss.}

% \noindent\textbf{Modality-wise IRM Loss.}

During the training stage, we formulate the overall training objective as: 
% a combination of the cross-entropy loss $\mathcal{L}_{C E}$, modality-wise IRM loss $\mathcal{L}_{inv}$ and the cross-modal alignment loss $\mathcal{L}_{cro}$ :
\begin{equation}
\label{eq:6}
\min _{G, E_{\text{2D}}, E_{\text{3D}} } \mathcal{L}_{C E}( E_{\text{2D}}, E_{\text{3D}})+ \mathcal{L}_{\text{inv}}(G) + \mathcal{L}_{\text{align}}(E_{\text{2D}}, E_{\text{3D}}),
\end{equation}
where $\mathcal{L}_{C E}$ is the standard cross-entropy loss, $\mathcal{L}_{\text{inv}}$ is the modality-wise IRM loss to optimize gate function $G$, and $\mathcal{L}_{\text{align}}$ is defined as follow~\footnote{Note that each loss optimizes different set of parameters --- the feature encoder $E_{\text{2D}}$ and $E_{\text{3D}}$ is frozen when IRM penalty updates; the gate function $G$ is only optimized by the modality-wise IRM loss.}: 

\noindent\textbf{Cross-modality Alignment Loss.} After the gate function $G$ filters the non-conflicting features, the multi-modality encoders ${E}_{\text{3D}}$, ${E}_{\text{2D}}$ are eventually regularized in collaborative feature space by the cross-modality NT-Xent loss~\cite{chen2020simple} without memory bank for further alignment:
\begin{equation}
\small
    \label{eq.7}
\mathcal{L}_{align} = -\log \frac{\exp \left(\boldsymbol{z}^{\mathrm{T}} \boldsymbol{z}^{+} \cdot \tau\right)}{\exp \left(\boldsymbol{z}^{\mathrm{T}} \boldsymbol{z}^{+} \cdot \tau\right)+\sum_{\boldsymbol{z}^{-} } \exp \left(\boldsymbol{z}^{\mathrm{T}} \boldsymbol{z}^{-} \cdot \tau\right)},
\end{equation}
where we use $\boldsymbol{z} = G(\mathbf{x}_{2})$ and $\boldsymbol{z}^{+} = G(\mathbf{x}_{3})$ for brevity; $\tau$ is a temperature parameter. The objective is to maximize the cosine similarity of $G( \mathbf{x}_{2})$ and $G(\mathbf{x}_{3})$, which are the 3D/2D non-conflicting feature of the same sample, while minimizing the similarity with all the others in the feature space for modality alignment.

% However, previous KD approaches usually assume that the training data used by the teacher and student are from the same distribution [17]. Thus, since sparse and disordered point clouds represent visual information different from images, naive feature alignment between two representations appeals to cause limited gains or negative transfer for the cross-modal scenario.

\subsection{ Joint Inference}
\label{sec:3.4}

% \noindent\textbf{Input} : 

% \noindent\textbf{Output} : Final joint-prediction.

We devise a simple multi-modality knowledge fusion strategy for joint prediction in inference. In Figure~\ref{fig:4}(c), the 3D branch $E_{\text{3D}}$ takes point clouds as input to predict classification logits $f_\text{3}(\mathbf{x}_3)$, and the 2D branch $E_{\text{2D}}$ takes multi-view images as input to produce a visual feature embedding $\mathbf{x}_{2}$ for each of them. To make the best of our 2D branch that initialized with the CLIP model, we follow pretext tasks of CLIP pretraining to use the cosine similarity of image-text pairs for logits computation. Specifically, we get the textual embedding $\mathbf{x}_\text{text}$ by placing category names into a pre-defined text template, \emph{e.g.}, \textit{``rendered point cloud of a big [CLASS]''} and feeding the filled template to the textual encoder of CLIP model. The image-text similarity for the $i$-th rendered image is computed as $\frac{\mathbf{x}_{\text {text }}^{\top}\cdot {\mathbf{x}_{2}^i}}{\left\|\mathbf{x}_{\text {text }}\right\|\left\|\mathbf{x}^{i}_{2}\right\|}$.
% \dnote{First describe how to generate textual embedding, and then tell the readers to compute the cosine similarity for predicting the 2D logits.} \cnote{The textual embedding $F_{text}$ is obtained by the pretrained CLIP texture encoder with handcraft textual prompts, which is constructed by placing category names into a pre-dined template. Then the visual-textual cosine similarity can be calculated by $\frac{\mathbf{F}_{\text {text }}^{\top}\cdot {{f}^\text{2d}_i}}{\left\|\mathrm{F}_{\text {text }}\right\|\left\|{f}_{i}^{2d}\right\|}$  }. 
Once we obtain the cosine similarity of each rendered image, we average them to get the classification logits $f_{\text{2}}(\bm{x}_2)$ from 2D branches. After that, the fused prediction is computed as
 \begin{equation}
 \label{eq.8}
f_{ens} = \text{Softmax}(f_{\text{2}}(\bm{x}_{2})/\varphi) \cdot \text{Softmax}({f_{3}}({\bm{x}_{3}})),
\end{equation}
where Softmax is leveraged to normalize the weight; $\varphi$ is served as a temperature modulator to calibrate the sharpness of 2D logits distribution. Through such simple logits fusion, $P_{ens}$ can effectively fuse the prior multi-modal knowledge and ameliorate few-shot point cloud classification.
% Settings:
% Classification Task: ModelNet10, ModelNet40, ShapeNet , ScanObjectNN, Toys4k ,ModelNet-C(鲁棒性测试）(appendix) full-dataset(appendix)； zero/few shot (paper)；

% 主实验2张表格；附实验 3个参考coop的图；

\section{Experiments}
\label{sec:4}

% % baseline : 3d pretrained models ; 2d pretrained models ; 3d+3d base models
% We conduct the following experiments on several downstream tasks with our proposed \textsc{InvJoint}: (a) 3D few-shot classification task on ModelNet10/40~\cite{wu20153d}, ScanObectNN~\cite{uy2019revisiting} and TOYS4K~\cite{stojanov2021using} (b) 3D classification task on ModelNet40~\cite{wu20153d} and Modelnet40-C~\cite{sun2022benchmarking} (c) 3D shape retrieval task on Modelnet40~\cite{wu20153d}, ShapeNet Core55~\cite{sfikas2017exploiting}.

\begin{figure*}[t]
  \begin{minipage}[b]{1.0\linewidth}
  \centerline{\includegraphics[width=188mm]{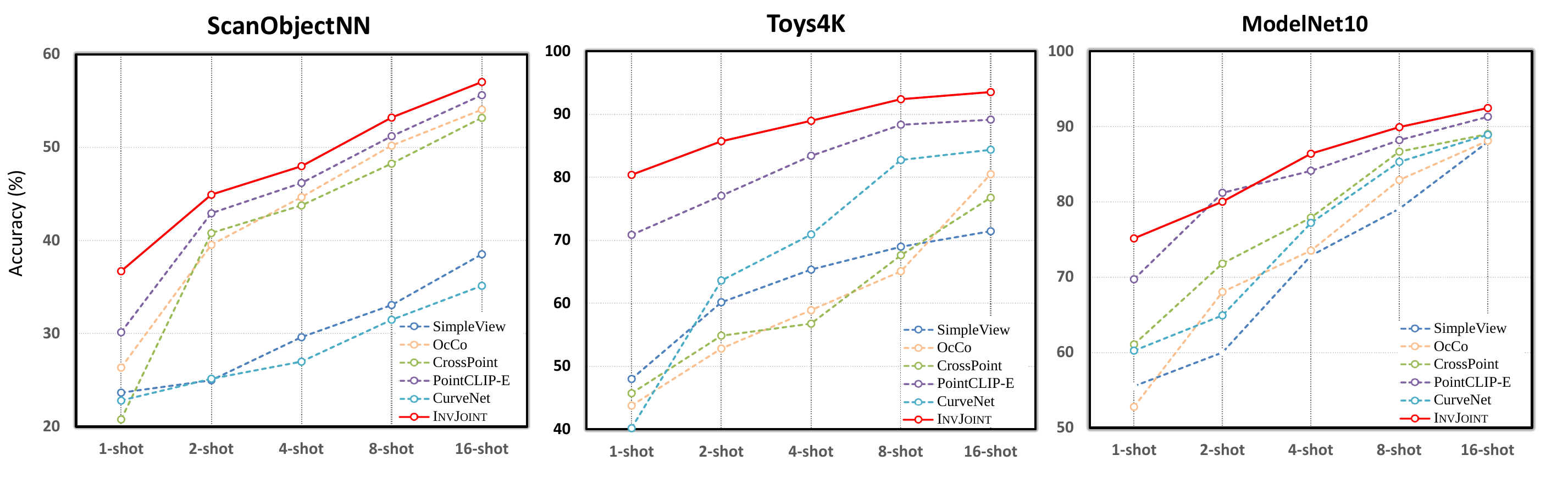}}
  \end{minipage}
  \caption{Few-shot performance comparisons between \textsc{InvJoint} and other methods, including the state-of-the-art PointCLIP-E ( denotes PointCLIP with simple late fusion), on ModelNet10, ScanObjectNN and Toys4K. Our \textsc{InvJoint} shows consistent superiority to other models under 1, 2, 4, 8, and 16-shot settings.}
  \label{fig:5}
\end{figure*}

\subsection{Implementation Details}

\noindent \textbf{Image Rendering.} We exploited a differentiable point cloud / mesh renderer (\emph{i.e.}, the alpha compositing / blending renderer~\cite{wiles2020synsin}). It uses learnable parameters  $\bm{r}=\left \{ \rho, \theta, \phi  \right \}$ to indicate the camera's pose and position, where $\rho$ is the distance to the rendered object, $\theta$ is the azimuth, and $\phi$ is the elevation. Other than the parameter $\bm{r}$, the light pointing is fixed towards the object center and the background is set as pure color. We further resized the rendered images to 224 $\times$ 224, and colored them by the values of their normal vectors or kept them white if normal is not available.

% \noindent \textbf{Image Rendering.} We exploit the differentiable point cloud renderer $R$~\dnote{Is there a specific name?} from PyTorch3D~\cite{ravi2020accelerating} in our pipeline for its speed and compatibility with PyTorch libraries~\cite{paszke2019pytorch}. We resize the size of each rendered image to 224 $\times$ 224, and color them by their normal's values or keep them white if their normal values are not available. An alpha compositing~\cite{wiles2020synsin} renderer is adopted with learnable parameters $\bm{r}=\left \{ \rho, \theta, \phi  \right \}$ to indicate the camera's pose and position, where $\rho$ is the distance to the rendered object, $\theta$ is the azimuth, and $\phi$ is the elevation. Other than the parameter $\bm{r}$, the light pointing is fixed towards the object center and the background is set as pure color. 

\noindent \textbf{Network Architectures.}
We adopted ViT-B/16~\cite{dosovitskiy2020image} and textual transformers pretrained with CLIP~\cite{radford2021learning} as our 2D visual encoder and textual encoder, respectively. Their parameters were frozen throughout our training stage.
% CLIP~\cite{radford2021learning} (with frozen visual \& textual encoder) as our 2D backbone. Specifically, we adopt the basic version of ViT-B/16~\cite{dosovitskiy2020image} as our visual encoder and transformers as the textual encoder by default. 
Following the practice in ~\cite{zhang2022pointclip}, we set the handcraft language expression template as `` 3D rendered photo of a [CLASS]" for textual encoding. As for 3D backbones, for fair comparison with other methods, we exploited the widely-adopted DGCNN~\cite{wang2019dynamic} point cloud feature encoder as ${E}_{\text{3D}}$. 

\noindent \textbf{Training Setup.}
\textsc{InvJoint} was end-to-end optimized at the training stage. For each point cloud input, we sampled 1,024 points via Farthest Point Sampling~\cite{moenning2003fast}, and applied standard data augmentations, including rotation, scaling, jittering, and color auto-contrast. For rendered images, we only applied center-crop as the data augmentation, since the background is purely white. \textsc{InvJoint} was trained for 50 epochs with a batch size of 32. We adopted SGD as the  optimizer~\cite{loshchilov2016sgdr}, and set the weight decay to $10^{-4}$ and the learning rate to 0.01. Cosine annealing~\cite{loshchilov2017decoupled} was employed as the learning rate scheduler. All of our experiments were conducted on a single NVIDIA Tesla A100 GPU.

% \textsc{InvJoint} is trained for 50 epochs with a batch size of 32. We adopt SGD optimizer~\cite{sgd} with weight decay 5e-4 and learning rate 0.01

\subsection{Few-shot Object Classification}

\noindent \textbf{Dataset and Settings.} We compared our \textsc{InvJoint} with other state-of-the-art models on four datasets: ModelNet10~\cite{wu20153d}, ModelNet40~\cite{wu20153d}, ScanObjectNN~\cite{uy2019revisiting} and Toys4K~\cite{stojanov2021using}. ModelNet40 is a synthetic CAD dataset, containing 12,331 objects from 40 categories, where the point clouds are obtained by sampling the 3D CAD models. ModelNet10 is the core indoor subset of ModelNet40, containing approximately 5k different shapes of 10 categories. ScanObjectNN is a more challenging real-world dataset, composed of 2,902 objects selected from ScanNet~\cite{dai2017scannet} and SceneNN~\cite{hua2016scenenn}, where the point cloud shapes are categorized into 15 classes. Toys4K~\cite{stojanov2021using} is a recently collected dataset specially designed for low-shot learning with 4,179 objects across 105 categories. 

We followed the settings in ~\cite{zhang2022pointclip} to conduct few-shot classification experiments: for $K$-shot settings, we randomly sampled $K$ point clouds from each category in the training dataset. Our point cloud encoder $E_{\text{3D}}$, as well as the other point-based methods in few-shot settings were all pretrained on ShapeNet~\cite{chang2015shapenet}, which originally consists of more than 50,000 CAD models from 55 categories.

\begin{table}[t]
\caption{Few-shot performance on ModelNet40 with several 2D/3D state-of-the-art methods. PointCLIP-E denotes the ensemble of PointCLIP and DGCNN.  }
\vspace{0.12cm}
\centering
\resizebox{0.48\textwidth}{!}{%
\begin{tabular}{c | c | c c c c c }
\toprule[1.4pt]
Category                      &  Method     &  1-shot  & 2-shot & 4-shot & 8-shot & 16-shot     \\ \midrule[1.2pt]
\multirow{2}{*}{2D}  &  PointCLIP~\cite{zhang2022pointclip}  &  52.96   & 66.73  & 74.47  & 80.96  & 85.45    \\ 
                              &  SimpleView~\cite{goyal2021revisiting} &  26.42   & 35.14  & 58.53  & 69.20  & 78.55  \\ \midrule
\multirow{3}{*}{3D}  &  OcCo~\cite{wang2021unsupervised}       &   46.92      & 54.08      & 60.15      & 72.98      & 75.08   \\  
                              &  cTree~\cite{sharma2020self}      &   15.13      & 24.98     & 27.90      & 34.12      & 50.59   \\
                              &  Jigsaw~\cite{sauder2019self}     &   11.24      & 20.98      & 25.76      &31.89      & 46.85   
                              \\ \midrule
\multirow{3}{*}{Joint}        &  Crosspoint~\cite{afham2022crosspoint} &  48.24   & 59.95  & 64.25  & 75.75  & 79.70   \\  
                              &  Shape-FEAT~\cite{stojanov2021using} & 37.78  & 49.92  & 54.10  &61.98   & 70.75   \\  
                              &  PointCLIP-E  &  53.70   & 67.14  & 76.32  & 80.82  & 85.90   \\  \midrule
Ours                          &  \textsc{InvJoint}       &  \textbf{68.85}   & \textbf{72.94}  & \textbf{78.95} & \textbf{83.61} &   \textbf{88.97}    \\ \bottomrule[1.4pt]

\end{tabular}}
\label{t1}
\end{table}

\noindent \textbf{Performance Comparison.} Table~\ref{t1} reports the few-shot classification performance on ModelNet40 dataset. Several state-of-the art methods, including image-based, point-based and multi-modality-based ones, are compared. Note that post-search in PointCLIP~\cite{zhang2022pointclip} is not leveraged for a fair comparison. 
Our \textsc{InvJoint} achieves inspiring performance, outperforming all other methods by a large margin. Remarkably, \textsc{InvJoint} achieves an absolute improvements of \textit{6.02} \% on average against PointCLIP~\cite{zhang2022pointclip}. The superiority of our method becomes more obvious when it comes to harder conditions with fewer samples. For example, in 1-shot settings, \textsc{InvJoint} outperforms PointCLIP~\cite{zhang2022pointclip} and Crosspoint~\cite{afham2022crosspoint} by \textit{15.89} \% and \textit{20.61} \% respectively. Besides, we can observe from Table~\ref{t1} that simply ensembling PointCLIP~\cite{zhang2022pointclip} and DGCNN~\cite{wang2019dynamic} couldn't provide enough enhancement, which demonstrates that the conventional ensemble strategy cannot work well without tackling the \textbf{``joint hard samples''}. 

Figure~\ref{fig:5} depicts the results in the other three datasets. Not surprisingly, \textsc{InvJoint} consistently outperforms other methods across datasets and in most settings, further demonstrating the robustness of our method. Particularly, in the recently collected benchmark Toys4K with the largest number of object categories, \textsc{InvJoint} shows an overwhelming performance, \emph{i.e.}, \textit{93.52} \% accuracy with 16-shots, while most  3D models achieve really low performance due to their poor generalization ability. 

% This suggests that the pre-trained knowledge from 2D modality is useful for solving 3D recognition problems and is better than directly pre-training on 3D datasets with low volume of data\dnote{Why talk about pretraining? your paper is a work for joint training and ensemble.}.

% From the results we can draw three conclusions.
% Although it has fewer categories than ModelNet40, it is more practically challenging than its synthetic counterpart due to complex backgrounds, missing parts, and various deformations.

% Firstly, our textsc{InvJoint} outperforms traditional 3D pretraining methods. This suggests that the pre-trained knowledge from 2D domain is useful for solving 3D recognition problems and is better than directly pre-training on 3D datasets with limited data.

% \cnote{add one sentence about pointclip ensemble, cue intro.}

% \cnote{Visualization of Joint hard samples}

% \noindent \textbf{Robustness}

% 3-lines Picture : ModelNet10;ScanObjectNN;Toys4k

% Table : ModelNet40

% Other exp : Modelnet40-C (Many-shot robustness)

% Appendix : ModelNet40 Many-shot 

% Other tasks (retrieval)

\subsection{Other Downstream Tasks}

Besides few-shot object classification, we also deployed \textsc{InvJoint} in the following downstream tasks to show its more collaborative 2D-3D features.

\noindent \textbf{Dataset and Settings.} We followed the settings in~\cite{hamdi2021mvtn} to provide the empirical results of 3D shape retrievals task on ModelNet40~\cite{wu20153d} and ShapeNet Core55~\cite{sfikas2017exploiting}. Furthermore, we also experienced \textsc{InvJoint} on ModelNet40 and ModelNet40-C for many-shot object classification. ModelNet40-C~\cite{sun2022benchmarking} is a comprehensive dataset with 15 corruption types and 5 severity levels to benchmark the corruption robustness of 3D point cloud recognition. Note that in all the three following downstream tasks, our point cloud encoder $E_{\text{3D}}$ is trained \textit{from scratch} for a fair comparison.

\begin{table}[]
\caption{Object classification results on ModelNet40 and Modelnet40-C.  ``Corr Err'' and ``Clean Err'' denote the error rate on ModelNet40-C and ModelNet40, respectively. }
% \centering
\resizebox{0.47\textwidth}{!}{%
\begin{tabular}{ c | c | c| c}
\toprule[1.4pt]
  Methods &   Augmentation  &  Corr Err  & Clean Err \\   \midrule[1.2pt]
   \multirow{2}{*}{PCT~\cite{guo2021pct}}           &    PointCutMix-K      & 16.5  & \textbf{6.9}  \\
                 &    PointCutMix-R      & 16.3   &7.2 \\  \midrule
    \multirow{2}{*}{DGCNN~\cite{wang2019dynamic}}         &    RSMix              & 18.1  &7.1 \\
         &    PointCutMix-R      & 17.3 &7.4\\  \midrule
    \multirow{2}{*}{PointNet++~\cite{qi2017pointnet++}}    &    PointCutMix-R      & 19.1  &7.1 \\
       &    PointMixup         & 19.3       &7.1   \\   \midrule
    SimpleView~\cite{goyal2021revisiting}    &    PointCutMix-R      & 19.7  &7.9 \\  \midrule
    RSCNN~\cite{hu2020rscnn}         &    PointCutMix-R      & 17.9   &7.6\\ \midrule
   {\textsc{InvJoint}} \scriptsize{(DGCNN)}        &    RSMix              & 16.8  \scriptsize{(\textcolor[rgb]{1,0,0}{1.3} \textcolor[rgb]{1,0,0}{$\downarrow $})}  & \textbf{6.9} \scriptsize{(\textcolor[rgb]{1,0,0}{0.2} \textcolor[rgb]{1,0,0}{$\downarrow $})}\\
    {\textsc{InvJoint}} \scriptsize{(PointNet++)}        &    PointCutMix-R               & 17.6  \scriptsize{(\textcolor[rgb]{1,0,0}{1.5} \textcolor[rgb]{1,0,0}{$\downarrow $})}  & 7.0 \scriptsize{(\textcolor[rgb]{1,0,0}{0.1} \textcolor[rgb]{1,0,0}{$\downarrow $})}\\
     {\textsc{InvJoint}} \scriptsize{(PCT)}        &    PointCutMix-K               & \textbf{15.9}  \scriptsize{(\textcolor[rgb]{1,0,0}{0.6} \textcolor[rgb]{1,0,0}{$\downarrow $})}  & \textbf{6.9} \scriptsize{(\textcolor[rgb]{1,0,0}{0.3} \textcolor[rgb]{1,0,0}{$\downarrow $})}\\

\bottomrule[1.4pt]
\end{tabular}}
\label{t3}
\end{table}

\begin{table}[]
\caption{3D Shape Retrieval. We compare the performance (mAP) of \textsc{InvJoint} on ModelNet40 and ShapeNet Core55. \textsc{InvJoint} achieves the best retrieval performance among recent state-of-the-art methods on both datasets.}
\vspace{0.1cm}
\centering
\resizebox{0.48\textwidth}{!}{%
\begin{tabular}{c | c | c| c }
\toprule[1.4pt]
  Methods      &      Data Type  & ModelNet40  & ShapeNet Core    \\ \midrule[1.2pt]
  PVNet~\cite{you2018pvnet}        &  Points   & 89.5    & -    \\ 
   Densepoint~\cite{liu2019densepoint}   &   Points  & 88.5     & -    \\
  RotNet~\cite{kanezaki2018rotationnet}        &  20 Views & -     & 77.2    \\ 
  MLVCNN~\cite{jiang2019mlvcnn}         &  24 Views  & 92.9        &     \\ 
  MVCNN~\cite{hamdi2021mvtn}        &  12 Views   & 80.2     & 73.5  \\ 
 MVTN~\cite{hamdi2021mvtn}         &  12 Views  & 92.2     & 82.9 \\ 
  ViewGCN~\cite{wei2020view}      &  20 Views & -     & 78.4 \\
 VointNet~\cite{hamdi2021voint}     & 12 Views &  -     & 83.3 \\ \midrule
  \textsc{InvJoint}   & 10 Views     & \textbf{93.7}     & \textbf{84.1}  \\ \bottomrule[1.4pt]
\end{tabular}}
\label{t2}
\end{table}

\noindent \textbf{(i) Shape Retrieval.}
For retrieval task, following~\cite{hamdi2021mvtn}, we leverage LFDA reduction~\cite{sugiyama2007dimensionality} to project and fuse the encoded feature (w/o the last layer for 3D branch) as the signature to describe a shape. Table~\ref{t2} presents the performance comparison with some recently introduced image-based and point-based methods in terms of mean average precision (mAP) for the shape retrieval task.
Note that some methods in Table~\ref{t2} are designed specifically for retrieval, \emph{e.g.}, MLVCNN~\cite{jiang2019mlvcnn}. Surprisingly, \textsc{InvJoint} improves the retrieval performance by a large margin in ShapeNet core with only 10 Views of rendered images. \textsc{InvJoint} also demonstrates state-of-the-art results (\textit{93.7} \% mAP) on ModelNet40. 

% Figure~\ref{fig:7} shows some qualitative  retrieval examples.

\noindent \textbf{(ii) Many-shot Object Classification}. Although our proposed \textsc{InvJoint} is mainly designed under few-shot settings, it can also achieve comparable performance with state-of-the-art methods on sufficient data. As depicted in Table~\ref{t3}, 
the performance of 3D baselines are significantly improved with lower error rate by \textsc{InvJoint}. Specifically, with PCT as the encoder $E_{3d}$ in 3D branch, we followed ~\cite{sun2022benchmarking} to conduct PointCutMix-K as point cloud augmentation strategy, our \textsc{InvJoint} achieve \textit{6.9} \% and \textit{15.9} \% error rate on ModelNet40 / ModelNet40-C respectively. 
% with DGCNN (\textit{from scratch}) as the encoder $E_{3d}$ in 3D branch, we followed ~\cite{sun2022benchmarking} to conduct PointCutMix-R / RsMix as point cloud augmentation strategy, our \textsc{InvJoint} surpasses most sophisticated methods and achieve \textit{6.9} \% error rate on ModelNet40. Moreover, it shows good robustness under data corruption, boosting the performance of DGCNN by \textit{1.3} \% in ModelNet40-C, attributing to the complementary and cooperative 2D-3D joint prediction.  

\begin{figure}[t]
   \begin{minipage}[b]{1.0\linewidth}
   \centerline{\includegraphics[scale = 0.34]{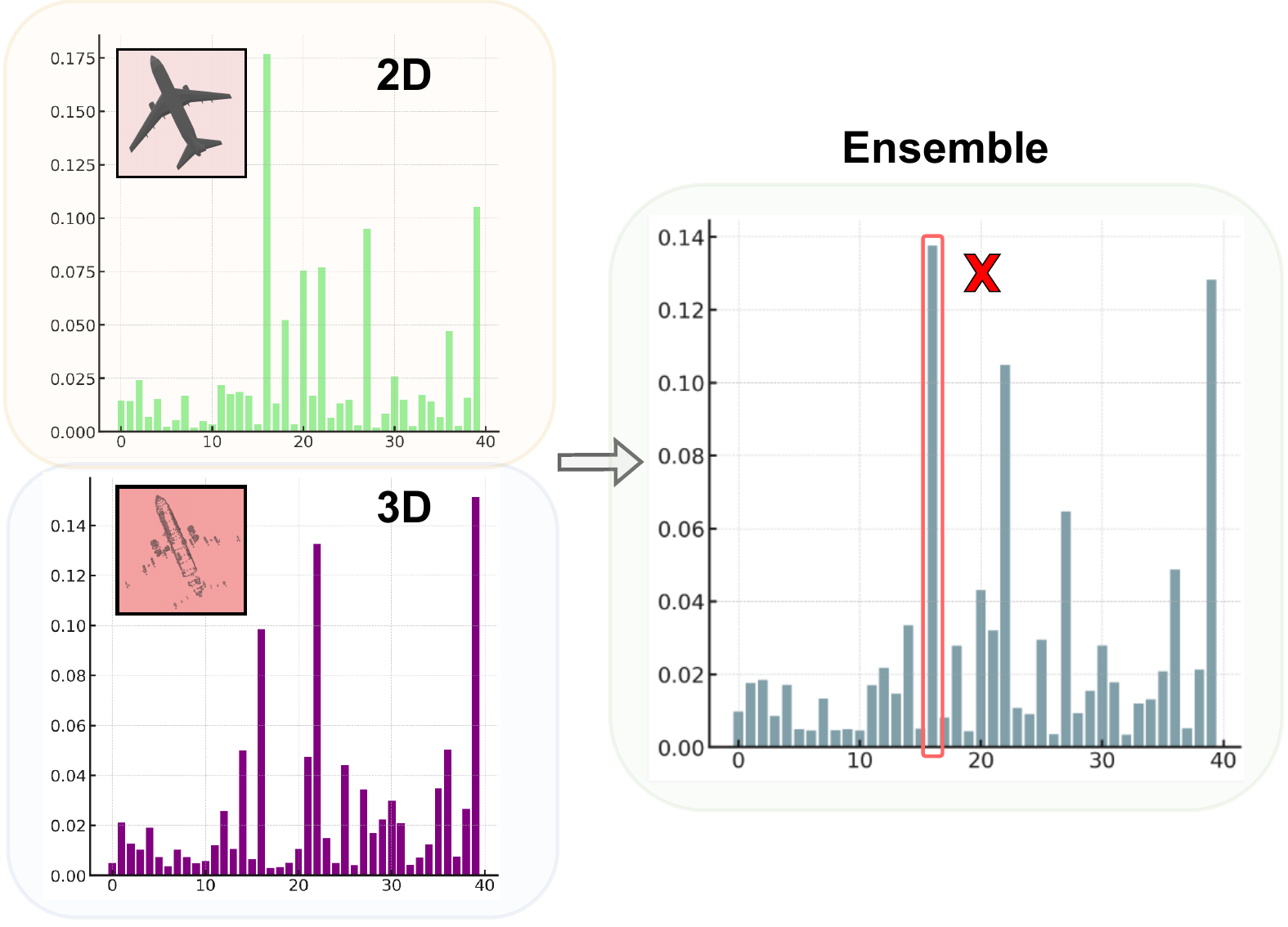}}
   \end{minipage}

\caption{The detailed failure cases caused by modality conflict. For a test sample with ground truth category \textbf{\textit{``39''}}, though 3D branch gives correct answer, the joint prediction is wrong because the 2D branch has high confidence on wrong category \textbf{\textit{``16''.}}}
   \label{fig:8}
\end{figure}

% \begin{table}[]
% \caption{Full Training-ModelNet40}
% \centering
% \resizebox{0.38\textwidth}{!}{%
% \begin{tabular}{c | c | c| c }
% \toprule[1.4pt]
% ID &   Pretrain  &  Methods      & Full-shot    \\ \midrule[1.2pt]
% 1  &    \multirow{4}{*}{N/A}    &  Pointnet++   & 90.7    \\
% 2  &  & PointMLP            & 94.1    \\ 
% 3  &  & PointNeXt            & 94.0   \\
% 4  &  & CurveNet             & 93.9  \\ \midrule
% 5  & \multirow{4}{*}{3D} & Dgcnn-ocCo          & 93.0  \\ 
% 6  &  &Ponit-BERT           & 93.2 \\ 
% 7  &  &Ponit-MAE             & 93.8 \\
% 8  &  & CrossPoint         & 91.2 \\ \midrule
% 9  & \multirow{3}{*}{2D}&  P2P              & 94.0  \\  
% 10 & & PonitCLIP            & 90.9 \\ 
% 11 & & INVJoint            & 93.8 \\
% \bottomrule[1.4pt]
% \end{tabular}}
% \label{t3}
% \end{table}

\subsection{Ablation Analysis}
\label{sec:4.4}
% (1) The component-wise contributions

% (2) The ratio of conflict

% (3) The robustness of backbone

% (4) Effectiveness: parameter and epochs

% (5) the success of ensemble (with oracle up-bound)

\begin{figure}[t]
   \begin{minipage}[b]{1.0\linewidth}
   \centerline{\includegraphics[width =9.1cm]{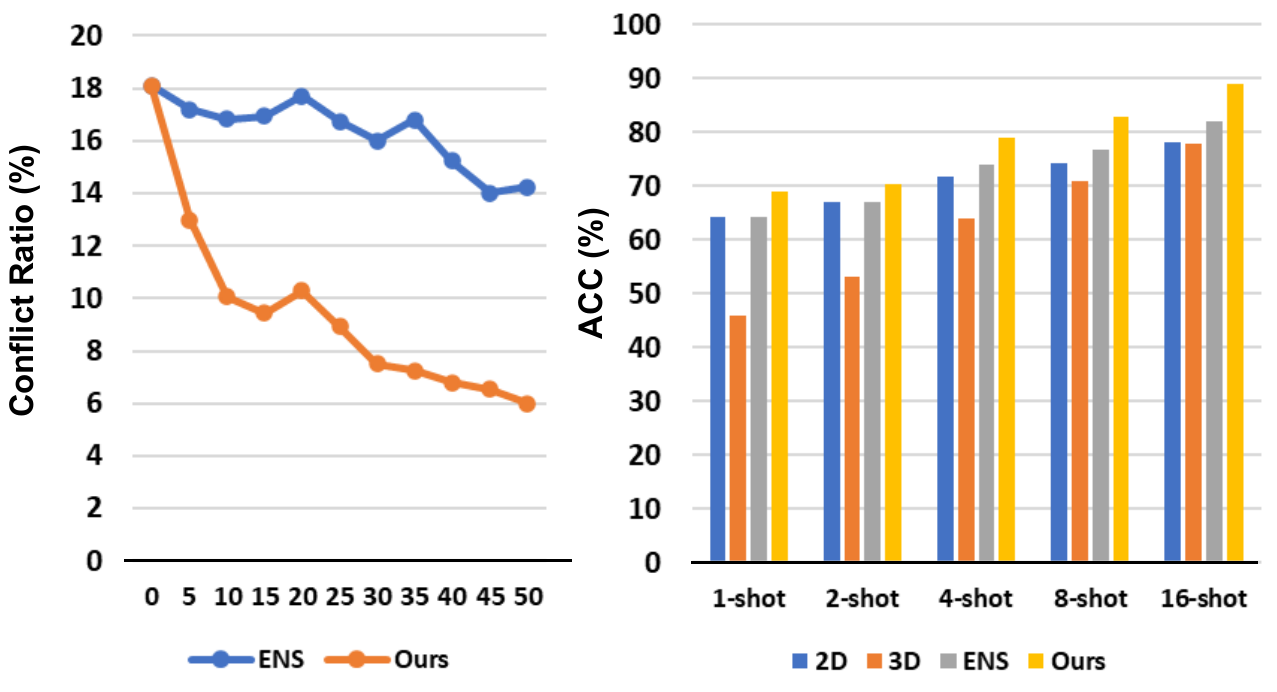}}
   \end{minipage}
   \vspace{0.05cm}
   \caption{ (a) The variation curve of the \textit{Conflict Ratio} $C_\text{err}$ on 16-shots ModelNet40,  which degrades significantly with \textsc{InvJoint}. (b)   Evaluations (Top-1 Accuracy) on ModelNet40 with different few-shot settings. Joint inference with \textsc{InvJoint} outperforms late fusion baseline by a large margin. }
   \label{fig:6}
\end{figure}

\noindent\textbf{Q1: }\emph{\textbf{How does \textsc{InvJoint} make the best of 2D and 3D world? }}
To better diagnose the effectiveness of \textsc{InvJoint}, we first illustrate the improvements of our joint inference, comparing with each branch performance as well as the simple late fusion in Figure~\ref{fig:6}(b). Then we give the decent definition of \textit{Conflict Ratio} $C_\text{err}$ to reflect the degree of modality conflict: Given the set of test sample index as $\mathbf{T}$, we define the index of samples with correct 2D, 3D and Joint predictions as $\mathbf{T_\text{2D}}$, $\mathbf{T_\text{3D}}$ and $\mathbf{T_\text{Joint}}$. $C_\text{err}$ is given by $\frac{||(\mathbf{T_\text{2D}} \setminus \mathbf{T}_\text{Joint}) \cup (\mathbf{T_\text{3D}} \setminus \mathbf{T}_\text{Joint})||}{||T||}$, which calculates the ratio of those can be recognized by one modality but failed in joint prediction
. Under such definition, we further analyze the variation curve of $C_\text{err}$ at the training stage.

\noindent\textbf{A1: } Specifically in Figure~\ref{fig:6}(b), the proposed \textsc{InvJoint} outperforms the late fusion by \textit{4.7} \% on average in different settings of ModelNet40, which concretely demonstrates the superiority of multi-modality collaboration through \textsc{InvJoint}. It is clear from Figure~\ref{fig:6}(a) that our method gradually mitigates the modality conflict while separate training of each branch and then ensembling remains high \textit{Conflict Ratio} $C_\text{err}$. Figure~\ref{fig:8} gives a detailed example for the failure of simple ensemble caused by modality conflict. From these two aspects, we could give a conclusion: the higher performance of \textsc{InvJoint} indeed attributes to the removal of conflict and ambiguous predictions.

\vspace{0.12cm}

\noindent\textbf{Q2: }\emph{\textbf{What impact performance of \textsc{InvJoint} considering component-wise contributions? }} we removed each individual step of Invariant Joint Training and replaced the late fusion strategy to examine the component-wise and loss-wise contributions. The results are shown in Table~\ref{t6}.

\noindent\textbf{A2: } In a multi-step framework, we observed that the exclusion of any component from \textsc{InvJoint} resulted in a significant decrease in performance. Specifically, upon removal of either step 1 or step 2, the Top-1 Accuracy exhibited an average degradation of \textit{4.06} \% \textcolor[rgb]{1,0,0}{$\downarrow $} and \textit{3.02} \% \textcolor[rgb]{1,0,0}{$\downarrow $}. Similarly considering the loss functions, the Top-1 Accuracy will averagely degrade by \textit{5.45 \%} \textcolor[rgb]{1,0,0}{$\downarrow $} and \textit{3.02 \%} \textcolor[rgb]{1,0,0}{$\downarrow $} respectively if $\mathcal{L}_{C E}$ is adopted alone (w/o Step1 \& 2) and if $\mathcal{L}_{\text{inv}}$ is not adopted (w/o Step 2). Further hyper-parameter sensitivity (\emph{e.g.}, \textbf{$\lambda$} in Eq~\eqref{eq:5}) analyses are included in \textit{Appendix}. 

% As for fusion strategy during \textit{inference}, both logits addition or multiplication could give great enhancement. In accordance with the empirical findings reported by  ~\cite{chen2022multimodal},
% logits multiplication outperforms its counterpart by \textit{0.87} \% on average. 

% We leave a theoretical discussion about the fusion type for multi-modality inference in \textit{Appendix}.
% The Top-1 Accuracy will averagely degrade by \textit{5.45 \%} \textcolor[rgb]{1,0,0}{$\downarrow $} and \textit{3.02 \%} \textcolor[rgb]{1,0,0}{$\downarrow $} respectively if $\mathcal{L}_{C E}$ is adopted alone (w/o Step1 \& 2) and if $\mathcal{L}_{\text{inv}}$ is not adopted (w/o Step 2).

% Specifically, upon removal of either step 1 or step 2, the Top-1 Accuracy exhibited an average degradation of 4.06% and 3.02%,

\begin{table}[]
\caption{Performances with different visual encoders on ModelNet40. RN50 /101 denotes ResNet-50 /101, and ViT-B/32 represents vision transformer with 32 × 32 / 16 × 16 patch embeddings. Accuracy of 2D branch (\textit{left in each cell}) and \textsc{InvJoint} (\textit{right in each cell}) are reported.}
\vspace{1mm}
\centering
\resizebox{0.49\textwidth}{!}{%
\begin{tabular}{c | c| c| c|c |c}
\toprule[1.4pt]
Model   &  1-shot      &   2-shot     & 4-shot      & 8-shot   &16-shot      \\ \midrule[1.2pt]
RN50  &    59.18 $\vert$ 66.05   &  64.12 $\vert$ 68.90      & 68.23 $\vert$ 76.41   & 71.10 $\vert$ 81.25 &  77.23 $\vert$ 85.93 \\
RN101  &   60.19 $\vert$ 66.42    &  \textbf{66.98} $\vert$ 70.31  & 70.45 $\vert$ 78.90   &     72.74 $\vert$ 82.60   & 78.16 $\vert$ 87.10\\ 
ViT/16   & \textbf{63.62} $\vert$ \textbf{68.85}      &   67.23 $\vert$ \textbf{72.94}    & \textbf{72.35} $\vert$ \textbf{78.95}  &    \textbf{75.68} $\vert$ \textbf{82.85}  &    \textbf{81.20} $\vert$ \textbf{88.97}\\
ViT/32 &  61.29 $\vert$ 67.34     &  66.08 $\vert$ 69.70     & 70.14 $\vert$ 77.62   &    73.14 $\vert$ 81.90   &  80.12 $\vert$ 88.32 \\\bottomrule[1.4pt]
\end{tabular}}
\label{t4}
\end{table}

\vspace{0.12cm}

\noindent\textbf{Q3: }\emph{\textbf{How about the robustness of \textsc{InvJoint}?}} As shown in Table~\ref{t4}, we compared the effect of different prompts designs for few-shot \textsc{InvJoint}. Moreover, we also implemented different CLIP visual backbones from ResNet~\cite{he2016deep} to ViT~\cite{dosovitskiy2020image}, reporting the results of individual 2D branch as well as the joint prediction of \textsc{InvJoint}.

\noindent\textbf{A3: } From Table~\ref{t4} and \ref{t5}, we could find out that the performance of 2D branch is directly impacted by the prompt and backbone choices to some extent. However, with the cooperative 3D-2D joint prediction, our proposed \textsc{InvJoint} shows its relatively strong robustness, \emph{e.g.}, reducing the standard deviation from \textit{10.87} \% to \textit{5.51} \% among the different designs of prompts. More empirical analysis on different point cloud augmentation strategies as well as the choices of 3D backbones is included in \textit{Appendix}. 

\begin{table}[]
\caption{Performances with different prompt designs on 16-shot Toys4K. [CLASS] denotes the class token, and [Learnable Tokens] denotes learnable prompts with fixed length.}
\vspace{0.1cm}
\centering
\resizebox{0.465\textwidth}{!}{%
\begin{tabular}{  c | c| c }
\toprule[1.4pt]
 Prompts &   $E_{2D}$  & Joint   \\ \midrule[1.2pt]
 ``a photo of a [CLASS].”                  &  88.18    & 92.90   \\
``a point cloud photo of a [CLASS].”      &  89.32    & 93.18   \\
``point cloud of a big [CLASS].”          &  89.71    & 92.95   \\
``3D CAD model of [CLASS].”                 &  90.10    & 93.33   \\
``3D rendered photo of [CLASS].”      &  89.14    & \textbf{93.52}   \\
``3D object of a big [CLASS].”         &  \textbf{90.32}    & 92.98   \\
``[Learnable Tokens] + [CLASS]”          &  60.76    & 78.57   \\
\bottomrule[1.4pt]
\end{tabular}}
\label{t5}
\end{table}

\begin{table}[]
\caption{Effectiveness for each component on few-shot ScanObjectNN and ModelNet40. Performance of 2-shot (\textit{left in each cell}) and 16-shot (\textit{right in each cell}) are reported.}
\vspace{0.1cm}
\centering
\resizebox{0.475\textwidth}{!}{%
\begin{tabular}{c  c c| c|c}
\toprule[1.4pt]
Step1 & Step2 & Fusion Type   &  ScanObjectNN     &   ModelNet40   \\ \midrule[1.2pt]

 \xmark & \xmark   &  $f_{2d} + f_{3d} $  & 39.13 $\vert$ 51.90 &  65.67 $\vert$  79.46\\
 \xmark & \xmark   &  $f_{2d} \times f_{3d} $  & 39.06 $\vert$ 52.21 &  66.14 $\vert$  81.09\\ \midrule
 \xmark & \cmark   &  $f_{2d} + f_{3d} $  & 41.90 $\vert$ 52.93 & 66.54 $\vert$  81.42  \\
 \xmark & \cmark   &  $f_{2d} \times f_{3d}$  &  39.94 $\vert$ 53.45 &  67.29 $\vert$  83.91\\ \midrule
 \cmark & \xmark   & $f_{2d} + f_{3d} $ & 41.60 $\vert$ 52.78 &  66.15 $\vert$ 84.71 \\
 \cmark & \xmark   & $f_{2d} \times f_{3d} $  & 42.72 $\vert$ 53.62 & 67.75 $\vert$ 85.82 \\ \midrule
 \cmark & \cmark   & $f_{2d} + f_{3d} $ & 44.16 $\vert$ 56.19 & 71.42 $\vert$ 87.61 \\
 \cmark & \cmark   & $f_{2d} \times f_{3d} $  & \textbf{44.91} $\vert$ \textbf{57.02} & \textbf{72.94} $\vert$ \textbf{88.97}
 \\\bottomrule[1.4pt]
\end{tabular}}
\label{t6}
\end{table}

\section{Discussion}
\label{sec:5}
% \noindent\textbf{Q4: }\emph{\textbf{ How to plug-and-play?}}
\noindent\textbf{Q1: }\emph{\textbf{Have any theoretical insights been provided on Joint Hard Samples?}} We consider joint hard samples from a probabilistic perspective based on Bayesian decomposition. 

\noindent\textbf{A1: }
Recent studies~\cite{yi2022identifying,tang2022invariant} mainly attribute classification failures to the contextual bias, which wrongly associate a class label with the dominant contexts in the training samples of the class. In our work, the context can be encoded in the 2D and 3D modality-specific features---thus we call it modality bias. Denote the modality-invariant class features $z_c$ and modality-specific features as $z_d$. The classification model $p(y=c|x)$ that predicts the probability of an image $x$ belonging to the class $y=c$ breaks down into:
\begin{equation}
    \label{eq.a}
p(y=c \mid z_{c},z_{d}) = p(y=c \mid z_{c}) \cdot \overbrace{\frac{p\left(z_{d} \mid y=c, z_{c}\right)}{p\left(z_{d} \mid z_{c}\right)}}^{\text{modality  bias}},
\end{equation}

In a large-scale dataset where the independent and identical distribution (IID) assumption is satisfied, each modalities could give robust classification regardless of the influence of modality bias. That is to say $z_{d}$ is independent of y,\emph{i.e.} $p\left(z_{d} \mid y=c, z_{c}\right)$ approaching $p(z_{d} \mid z_{c})$, thus the attribute bias could be considered as constant. However, it is not always the case with data efficiency, where with different distribution of $z_{d}$ among modalities, resulting in different influence of modality bias.

\noindent\textbf{Joint Hard Samples}. From Eq.~\eqref{eq.a}, the modality bias  will largely decrease the performance if ${\exists} r \in \left \{ r_{2D}, r_{3D}  \right \}, r \neq c, p\left(z_{d} \mid y=r, z_{c}\right)  > p\left(z_{d} \mid y=c, z_{c}\right)$, making the sample hard to identify in both modalities; the diverse distribution shift of $z_{d}$ between 2D and 3D, making them confused on different sparks, denoted as $r_{2D} \neq r_{3D}$, giving high-confident prediction on different categories. 
% \dnote{Not clear yet. Even haven't mentioned joint head samples in this paragraph.}

% \dnote{First define what is joint hard example, and then explain joint example with Eq. 1.}

% \cnote{Why we need to select joint hard samples}
% Since the previous modality bias in \cite{yi2022identifying,tang2022invariant} is revealed in the same model, it hurts the OOD generalization and decreases the performance under distribution shifts. However, the modality bias in the proposed 2D-3D ensemble module is multi-modal and come from different networks, \ie, the second term in Eq.~\eqref{eq.1} is only variant across models. In fact, for a specific modality, some descriptive feature $z_d$ is even beneficial. For example, when $p_{2D}\left(z_{d} \mid y=c, z_{c}\right)$ is significantly larger than $p_{2D}\left(z_{d} \mid y \neq c, z_{c}\right)$ for 2D images from class $y=c$, such an modality-specific $z_d$ is good for 2D classification even if it's not hold in 3D point clouds, \eg, fine-grained texture in 2D images disappear in 3D representation. Therefore, with an ensemble paradigm, instead of removing all the modality bias as \cite{yi2022identifying,tang2022invariant} did, we need to find (reweight) joint hard samples and only removing the conflict while keeping some beneficial $z_{d}$ for each modalities. 

\begin{figure}[t]
   \begin{minipage}[b]{1.0\linewidth}
   \centerline{\includegraphics[scale = 0.49]{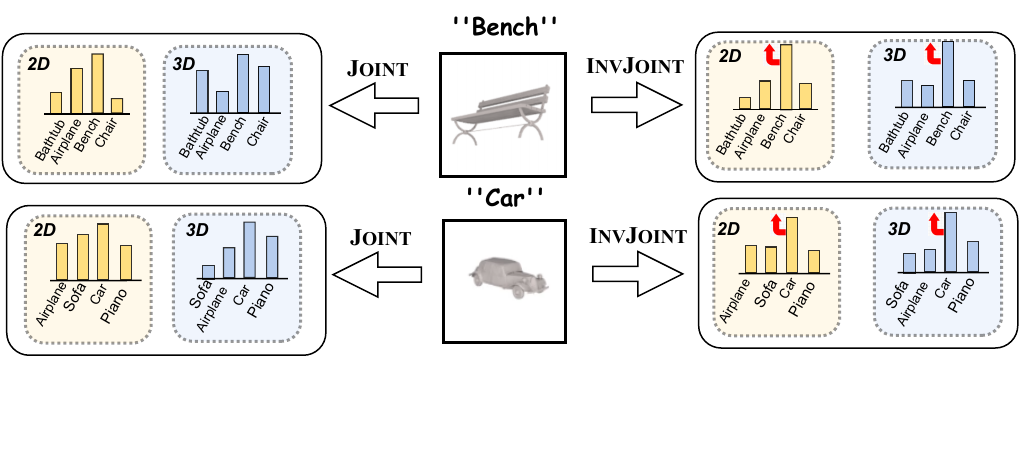}}
   \end{minipage}
\caption{The logit distribution of \textbf{joint hard samples} with/without \textsc{InvJoint}.}
   \label{fig:a}
\end{figure}

Additionally, It shows in Eq.~\eqref{eq.a} that the classification made by different modalities is biased due to the different sparks of modality bias $\frac{p\left(z_{d} \mid y=c, z_{c}\right) }{p\left(z_{d} \mid z_{c}\right)}$, which causes conflicting predictions especially under insufficient data scenarios.
Intuitively, the crux to mitigating such bias is to directly eliminate the impact of certain modality-specific $z_{d}$ distributions. Therefore, we treat the 2D/3D branches as two distinct learning environments, ensuring diverse $\frac{p\left(z_{d} \mid y=c, z_{c}\right)}{p\left(z_{d} \mid z_{c}\right)}$ across environments. Then, IRM essentially regularizes the invariant feature selector $G$ to achieve equivalent optimality across environments via the gradient penalty term in eq.(3). As a result, the influence of modality bias is eliminated, leading to the acquisition of a non-conflicting feature space $G(\mathbf{x}_{e})$ for further cross-modality alignment. Fig.~\ref{fig:a} illustrates that (1) joint hard samples differ in different spikes. (2) \textsc{InvJoint} removes the ambiguous predictions for a better ensemble. Due to limited space, please refers to ~\textit{Appendix} for further discussion.

\noindent\textbf{Q2: }\emph{\textbf{ Why Ensemble?}} One may ask why multi-modality ensembling should be regarded as an interesting contribution, since ensembling itself is a well-studied approach~\cite{webb2020ensemble,guptaensembles} that is often viewed as an  ``engineering stragety'' for improving leader board performance.

\noindent\textbf{A2: }  We would like to justify: \textbf{(1)} We illustrate that ensemble without conflict matters, and prior 3D-2D approaches such as knowledge distillation~\cite{yan2022let}, parameter inflation~\cite{xu2022image2point} are not as effective as \textsc{InVJoint}, especially under data deficiency. \textbf{(2)} As far as we know, \textsc{InVJoint} is the first 3D-2D ensembling framework as a fusion method for few-shot pointcloud recognition. What we propose is neither the added 2D classifier (a necessary engineering implementation) nor the ensemble paradigm in Figure~\ref{fig:1}(c), but a joint learning algorithm to improve the ineffective 2D-3D ensemble. Though simple, it is remarkably useful and should be considered as a strong baseline for future study.
\section{Conclusion}
We pointed out the crux to a better 2D-3D ensemble in few-shot point cloud recognition is the effective training on ``joint hard samples'', which implies the conflict and ambiguous predictions between modalities. To resolve such modality conflict, we presented \textsc{InvJoint}, a plug-and-play training module for ``joint hard samples'', which seeks the invariance between modalities to learn more collaborative 3D and 2D representation. Extensive experiments on 3D few-shot recognition and shape retrieval datasets verified the effectiveness of our methods. In future, we will focus on exploring the potential of \textsc{InvJoint} for wider 3D applications, \emph{e.g.}, point cloud part segmentation, object detection.

\section*{Acknowledgments}
This research is supported by the National Research Foundation, Singapore under its AI Singapore Programme (AISG Award No: AISG2-RP-2021-022) and the Agency for Science, Technology AND Research (A*STAR).

 \clearpage 
 \appendix

\noindent The \textbf{Appendix} is organized as follows:
\begin{itemize}[leftmargin=*]
    \item \textbf{Section~\ref{a}:} provides more details about our training pipeline. Specifically, we detailed the implementation of 2D renderer, CLIP linear adapter as well as the modality fusion and invariant risk minimization (IRM).
    \item \textbf{Section~\ref{b}:} gives further discussion on joint hard sample from probability theory and Venn graph.
    \item \textbf{Section~\ref{c}:} shows more experiment results and ablation studies, \eg, augmentations, OHEM strategy and parameter sensitivity analysis.
\end{itemize}

\section{Implementation Details}
\label{a}
As for 2D branch, We leverage the pretrained 2D knowledge for better point cloud analysis from two perspectives. 1) Beyond directly conducting CLIP visual encoder to projected depth maps in previous settings~\cite{zhang2022pointclip,goyal2021revisiting}, with the guidance of the frozen pretrained weights, the inputs of this branch can extensively bridge the modality gap between regular ones in 2D pretrained datasets and those transformed from point clouds through differentiable renderer. 2) Since fine-tuning the whole CLIP visual backbone would easily result in over-fitting under few-shot settings, we followed the strategy of PointCLIP~\cite{zhang2022pointclip}, freezing CLIP's visual and textual encoders and optimize the lightweight bottleneck adapter with the cross-entropy loss.

\noindent \textbf{Differentiable Renderer.} 
The renderer R, grounded in alpha compositing~\cite{wiles2020synsin}, is tasked with generating a rasterized object interpretation, utilizing the provided camera parameters. Learnable parameters $r=\left \{ \rho, \theta, \phi  \right \}$ are specifically harnessed to illustrate the camera's pose and position, with $\rho$ representing the distance to the object rendered, $\theta$ embodying the azimuth, and $\phi$ denoting the elevation. We employ a differentiable renderer to optimize the generation of pseudo images for improved recognition, which parameter is established through the confidence correlation between ground-truth label-generated prompt and the zero-shot performance of CLIP on downstream training datasets. For ModelNet40, Toys4K, and ShapeNet-Core, we utilize a differentiable mesh renderer. We maintain a fixed light source directed towards the object's center, applying normal vectors for coloration, or default to white when these are not accessible. As for ScanObjectNN, we deploy a differentiable point cloud renderer with 2048 points. This serves as a lighter alternative to mesh rendering in instances where CAD is unavailable or the mesh contains a significant number of obstructive faces~\cite{hamdi2021mvtn}.

\begin{figure}[t]
   \begin{minipage}[b]{1.0\linewidth}
   \centerline{\includegraphics[scale = 0.64]{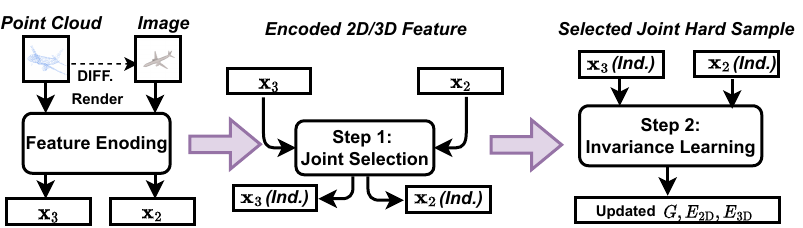}}
   \end{minipage}
   \caption {The simplified training framework.}
\end{figure}

\begin{table}[t]
  \centering
  \resizebox{0.48\textwidth}{!}{
    \begin{tabular}{ll|ll}
      \rowcolor{lightgray!30}
      Symbol & Definition & Symbol & Definition \\
      $\mathbf{x}_{2}$ / $\mathbf{x}_{3}$ & Encoded 2D/3D features  & $y$ & Class label  \\
      $\boldsymbol{z}_{e}$ &  Masked feature in each modality & $r$ & Threshold parameter \\
      $G$                  & Gate function     & $\lambda$   & IRM penalty weight \\
      $\theta$   & Dummy classifier, calculating gradients & $E_{\text{2D}}/E_{\text{3D}}$  & 2D/3D Feature encoder \\
      
    \end{tabular}
  }
  % \captionsetup{font={scriptsize}} 
  \vspace{-0.20cm}
  \caption{Notation Table}
  \label{tab2}
\end{table}

\noindent \textbf{Multi-view Feature Encoding. } We simplify the inter-view adapter for PointCLIP to further encode the N-view image feature $F_{I}$ with a proposed Multi-View adapter, which could capture the global and weighted view-wise feature simultaneously. With such simplification, we reduce the learnable parameters and avoid \textit{post-search}. 
 Specifically, given the N-view grid features $\mathbf{F}_{v}^{D}$, we first concatenate along the channel to obtain the global feature, then an aggregation function $A(\cdot)$ is calculated based on the pairwise affinity matrix $T \in \mathbb{R}^{N \times N}$ with feature cosine similarity. By aggregating the view-level vectors via $A(\cdot)$, we integrates shape information by reweighted view-wise feature representation. Finally, the encoding process can be formulated as:
\begin{equation}
\mathcal{F}_{\text{Global}} = f_{2}\left(\operatorname{ReLU}\left(f_{1}\left(\operatorname{concat}\left(\{\mathbf{F}_{v}^{D}\}_{v=1}^{N}\right)\right)\right)\right)
\end{equation}

\begin{equation}
\mathcal{F}_{\text{View}}=\operatorname{ReLU}\left(A\left(\operatorname{concat}\left(\{\mathbf{F}_{v}^{D}\}_{v=1}^{N}\right)\right)\right)
\end{equation}

\begin{equation}
\mathcal{F}_{I} = (1-\delta) \mathcal{F}_{\text{Global}} + \delta \mathcal{F}_{\text{View}}
\end{equation}

, where $A(\cdot)$ is the reweighting function, $f_{1}$ and $f_{2}$ are two-layer MLPs, and $\delta$ is mix-up combination coefficient.

% \noindent \textbf{2.5D Environment Construction.} 

% \noindent \textbf{Gradient Adjustment. }

\noindent \textbf{Modality Fusion.} As an \textit{optional} enhancement, we introduce a bidirectional attention mechanism, bridging the inherent strengths of 2D and 3D modalities to birth an intermediate \textbf{\textit{2.5D}} representation. This meticulously crafted modality proficiently harnesses localized nuances from both 2D images and 3D point clouds, imbibing the complementary details unique to each domain.

In detailed, our architecture is enriched with a bidirectional cross-attention layer~\cite{vaswani2017attention}. In the forward direction, point cloud features emerge as the query tensor, with the image features serving the dual roles of key and value tensors. In stark contrast, the reverse direction sees the image features donning the mantle of the query, whilst the point cloud features settle as both the key and value tensors. This duality in approach guarantees a harmonious balance in the weighing of features, rooted in mutual affinities. The culmination is a fusion that harmoniously encapsulates the distinctiveness of both modalities: 
\vspace{-1mm}
\begin{align}
Q_{\mathbf{x}_{3}} &= \mathbf{x}_{3}W_{Q}, & K_{\mathbf{x}_{2}} &= \mathbf{x}_{2}W_{K}, \\
V_{\mathbf{x}_{2}} &= \mathbf{x}_{2}W_{V}, & X_{\text{fused}} &= \text{softmax}(Q_{\mathbf{x}_{3}}K_{\mathbf{x}_{2}}^T) V_{\mathbf{x}_{2}}, \\
Q_{\mathbf{x}_{2}} &= \mathbf{x}_{2}W_{Q^{'}}, & K_{\mathbf{x}_{3}} &= \mathbf{x}_{3}W_{K^{'}}, \\
V_{\mathbf{x}_{3}} &= \mathbf{x}_{3}W_{V^{'}}, & X_{\text{fused-I}} &= \text{softmax}(Q_{\mathbf{x}_{2}}K_{\mathbf{x}_{3}}^T) V_{\mathbf{x}_{3}}.
\end{align}

The subsequent melding of \(X_{\text{fused}}\) and \(X_{\text{fused-I}}\) births an enriched feature set, \(X_{\text{bi-fused}}\), echoing the attributes of both 2D images and 3D point clouds. Note that such fusion-based modality is only served as a optional regularization term for calculating modality-wise IRM loss. Therefore, there is no additional classification head, and thus not leveraged during inference for the \textbf{\textit{2.5D}} modality environment.

\begin{figure}[t]
   \begin{minipage}[b]{1.0\linewidth}
   \centerline{\includegraphics[scale = 0.32]{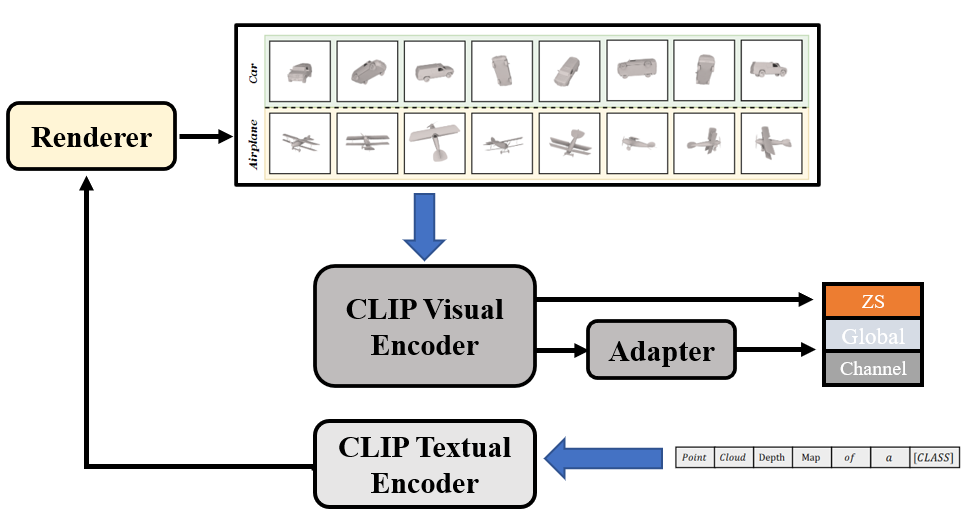}}
   \end{minipage}
   \caption {Detailed structure of 2D branch.}
   \label{fig:e}
\end{figure}

\noindent \textbf{Advanced IRM.}
In the manuscript, we introduced the modality-wise IRM to capture the common feature for better alignment. For its practical implementation~\footnote{We discards the dummy classifier $w$ and calculate Min-Max or variance of risks as the penalty term of IRM.}, we transitioned to REx~\cite{krueger2021out}, an optimized version especially adept under co-variate shifts. The MM(Min-Max)-REx is given by:
\begin{equation}
\begin{aligned}
\text{MM-REx}(\theta) &= \max_{\substack{\sum_{e=1}^m \lambda_e = 1\\ \lambda_e \geq \lambda_{\text{min}}}} \sum_{e=1}^{m} \lambda_e \mathcal{L}_e(\theta) \\
&= (1 - m\lambda_{\text{min}}) \max_{e} \mathcal{L}_e(\theta) + \lambda_{\text{min}} \sum_{e=1}^{m} \mathcal{L}_e(\theta).
\end{aligned}
\end{equation}

With goals akin to IRM, REx ensures invariance across environments in a more efficient and stable manner. We also leveraged the V-REx variant with additional 2.5D modality:
\begin{equation}
\text{RV-REx}(\theta) = \beta \mathrm{Var}(\{\mathcal{L}_1(\theta), ..., \mathcal{L}_m(\theta)\}) + \sum_{e=1}^m \mathcal{L}_e(\theta).
\end{equation}
Here, $\beta$ regulates between reducing average risk and ensuring risk consistency. Specifically, with $\beta = 0$ it aligns with ERM, while a higher $\beta$ emphasizes risk equalization.
Specifically, our modality-wise IRM differs from traditional ones as we utilize contrastive objectives $\mathcal{L}_{e}$, which show high efficiency for learning discriminative features. After filtering out conflicting features, we finally regularize $E_{\text{3D}}$, $E_{\text{2D}}$ in the collaborative feature space $G(\mathbf{x}_{e})$, aligning them with the cross-modality NT-Xent loss $\mathcal{L}_{align}$. 

\begin{figure}[t]
   \begin{minipage}[b]{1.0\linewidth}
   \centerline{\includegraphics[width =8.4cm]{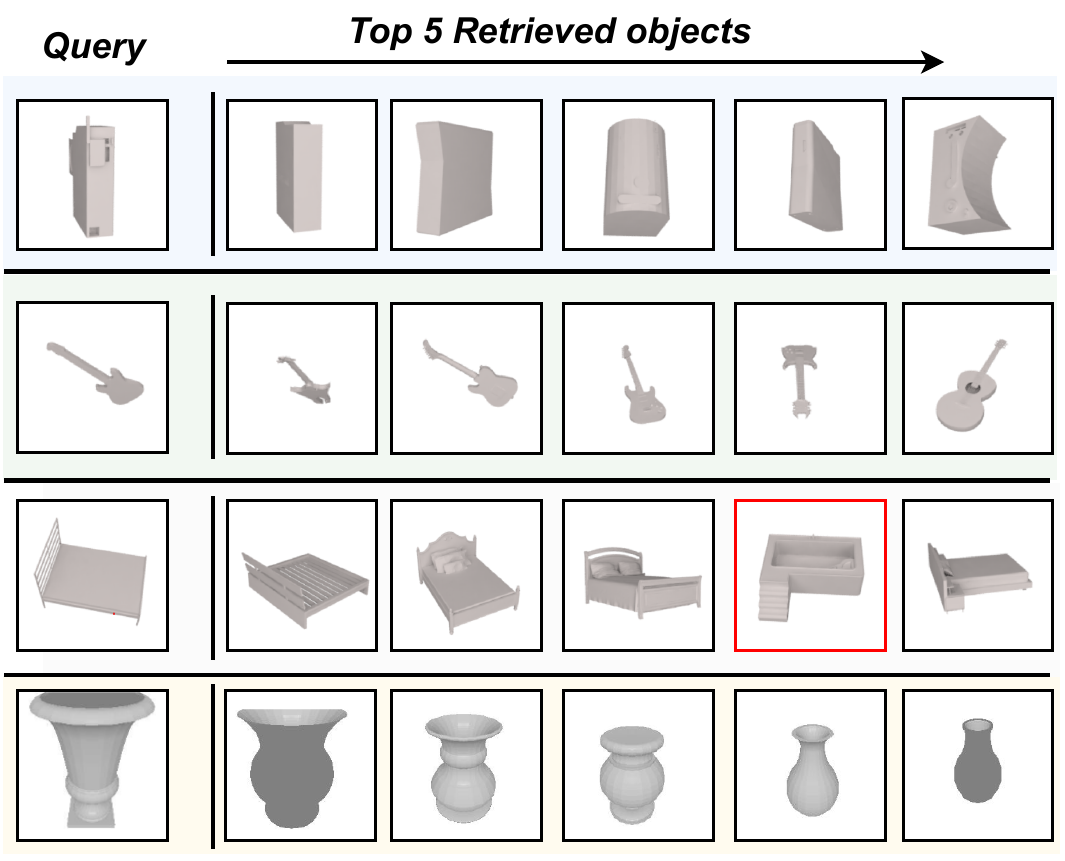}}
   \end{minipage}
   \caption{Qualitative Examples for 3D Shape Retrieval on ModelNet40: (\textit{left}): Query objects from the test set. (\textit{right}): Top 5 matches +for each query, with mistakes highlighted in red.}
   \label{fig:g}
\end{figure}

\noindent \textbf{Object Retrieval.} we leverage LFDA reduction to project and fuse the encoded feature (w/o the last layer for 3D branch) as the signature to describe a shape, which is further compared through Kd-Tree searching. Figure~\ref{fig:g} shows some qualitative  retrieval examples.

% \noindent \textbf{Logit Calibration.}

\section{Discussion On Joint Hard Sample }
\label{b}
\noindent \textbf{Probability Theory Perspective}. We still refer to the Bayesian Decomposition introduced in the manuscript:
\begin{equation}
    \label{eq.b}
p(y=c \mid z_{c},z_{d}) = p(y=c \mid z_{c}) \cdot \overbrace{\frac{p\left(z_{d} \mid y=c, z_{c}\right)}{p\left(z_{d} \mid z_{c}\right)}}^{\text{modality  bias}},
\end{equation}
Since the previous modality bias in \cite{yi2022identifying,tang2022invariant} is revealed in the same model, it hurts the OOD generalization and decreases the performance under distribution shifts. However, the modality bias in the proposed 2D-3D ensemble module is multi-modal and come from different networks, \ie, the second term in Eq.~\eqref{eq.b} is only variant across models. In fact, for a specific modality, some descriptive feature $z_d$ is even beneficial. For example, when $p_{2D}\left(z_{d} \mid y=c, z_{c}\right)$ is significantly larger than $p_{2D}\left(z_{d} \mid y \neq c, z_{c}\right)$ for 2D images from class $y=c$, such an modality-specific $z_d$ is good for 2D classification even if it's not hold in 3D point clouds, \eg, fine-grained texture in 2D images disappear in 3D representation. Therefore, considering ensemble, instead of removing all the modality bias as \cite{yi2022identifying,tang2022invariant} did, we need to find (reweight) joint hard samples and only removing the conflict while keeping some beneficial $z_{d}$ within modalities.

\begin{figure}[t]
   \begin{minipage}[b]{1.0\linewidth}
   \centerline{\includegraphics[scale = 0.64]{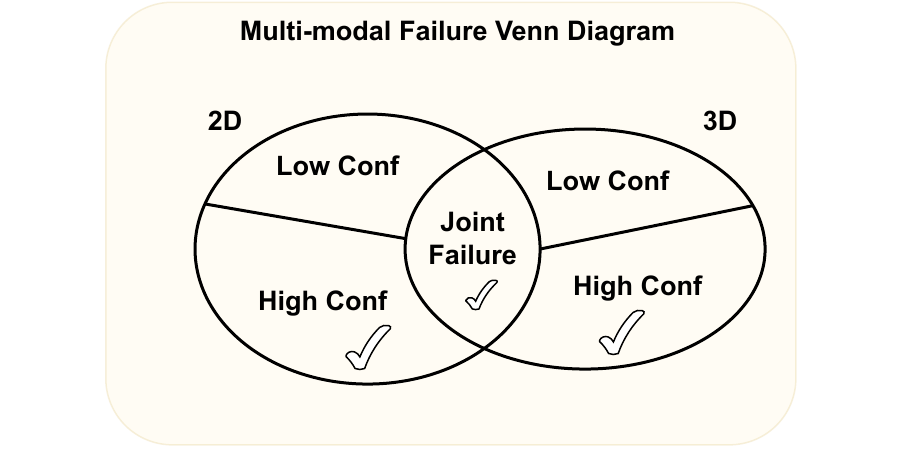}}
   \end{minipage}
   \caption  { The diagram of multi-modal failure Venn, where \textbf{Conf} denotes model's confidence on certain category for a test sample.   }
   \label{fig:b}
\end{figure}

\noindent \textbf{A Venn Diagram Perspective}. We maintain the assumption that, under few-shot fine-tuning, the primary improvement from the ensemble paradigm arises from the reduction of conflicting predictions, rather than from enhancements of a specific modality~\footnote{In other words, empirical findings suggest that joint 2D-3D training, without resorting to either contrastive or distillation methods, does not enhance the performance of a particular modality when compared to individual training in \textbf{few-shot settings}.}. Figure~\ref{fig:b} illustrates that the essence of an effective 2D-3D ensemble lies in minimizing the high confidence assigned to incorrect labels. In this pursuit, our invariant training strategy is anchored on ensuring the consistency of different (2D-3D) representations, serving to curtail biases stemming from conflicting modalities. An alternative method involves calibrating~\cite{kumar2022calibrated} the 2D/3D logits within each modality, balancing between confidence \textit{\textbf{avg.}} and accuracy \textit{\textbf{avg.}}. However, this technique demands a robust validation set, which is a rarity in the current 3D standard datasets. Looking ahead, we intend to juxtapose and adapt our invariant learning approach with the logit calibration model, especially within the context of the latest large-scale 3D datasets~\cite{wu2023omniobject3d,yu2023mvimgnet}.

% \subsection{Preliminary} 

\begin{table}[h]
\caption{Full Training on ModelNet40 with regular setting.}
\centering
\resizebox{0.38\textwidth}{!}{%
\begin{tabular}{c | c | c| c }
\toprule[1.4pt]
ID &   Pretrain  &  Methods      & Full-shot    \\ \midrule[1.2pt]
1  &    \multirow{4}{*}{N/A}    &  PointNet++~\cite{qi2017pointnet++}   & 90.7    \\
2  &  & PointMLP~\cite{ma2022rethinking}            & \textbf{94.1}    \\ 
3  &  & PointNeXt~\cite{qian2022pointnext}            & 94.0   \\
4  &  & CurveNet~\cite{muzahid2020curvenet}             & 93.9  \\ \midrule
5  & \multirow{4}{*}{3D} & Dgcnn-ocCo~\cite{wang2021unsupervised}          & 93.0  \\ 
6  &  &Ponit-BERT~\cite{yu2022point}           & 93.2 \\ 
7  &  &Point-MAE~\cite{pang2022masked}             & 93.8 \\
8  &  & CrossPoint~\cite{afham2022crosspoint}         & 91.2 \\ \midrule
9  & \multirow{3}{*}{2D}&  P2P~\cite{wang2022p2p}              & 94.0  \\  
10 & & PonitCLIP~\cite{zhang2022pointclip}            & 90.9 \\ 
11 & & INVJoint            & 93.9 \\
\bottomrule[1.4pt]
\end{tabular}}
\label{tp1}
\end{table}

\section{More Experiment}
\label{c}
\subsection{Additional Results on Many-shot ModelNet40} 
In the main manuscript, we followed ~\cite{sun2022benchmarking} and presented its performance for comparison. It is interesting to discuss the results under the regular settings as ~\cite{goyal2021revisiting}, thus we compared \textsc{InVJoint} with previous state-of-the-art methods under the full training ModelNet40. Note that in this setting, we use point cloud rendering instead of mesh rendering for data pre-processing. From Table~\ref{tp1}, our method shows comparable results with state-of-the-art methods on sufficient data. This further supports that our proposed framework can adapt to many-shot settings.

\subsection{Additional Results on Data Efficient Learning}
We follow ~\cite{yan2022let}, evaluating \textsc{InvJoint} under limited data scenario. From Table~\ref{tp5}, we could find that \textsc{InvJoint} consistently retains robust performance and out-performs PointCMT in all cases.

\begin{table}[]
\caption{Data efficient learning on ModelNet40.}
\vspace{0.1cm}
\centering
\resizebox{0.43\textwidth}{!}{%
\begin{tabular}{  c | c| c }
\toprule[1.4pt]
Data percentage &   w/ PointCMT~\cite{yan2022let}  & \textsc{InvJoint}   \\ \midrule[1.2pt]
 2 \%                  &  75.2   & 79.4 (\textcolor[rgb]{0,0,1}{+ 4.2} ) \\
 5\%      &  83.5   &  85.4  (\textcolor[rgb]{0,0,1}{+ 1.9} )  \\
10 \%         &  87.9    &  89.7 (\textcolor[rgb]{0,0,1}{+ 1.8} )   \\
20 \%                 &  89.3    & 91.3 (\textcolor[rgb]{0,0,1}{+ 2.0} ) \\

\bottomrule[1.4pt]
\end{tabular}}
\label{tp5}
\end{table}

\subsection{Additional Results on Augmentation Strategies} 
\begin{figure}[t]
   \begin{minipage}[b]{1.0\linewidth}
   \centerline{\includegraphics[scale = 0.40]{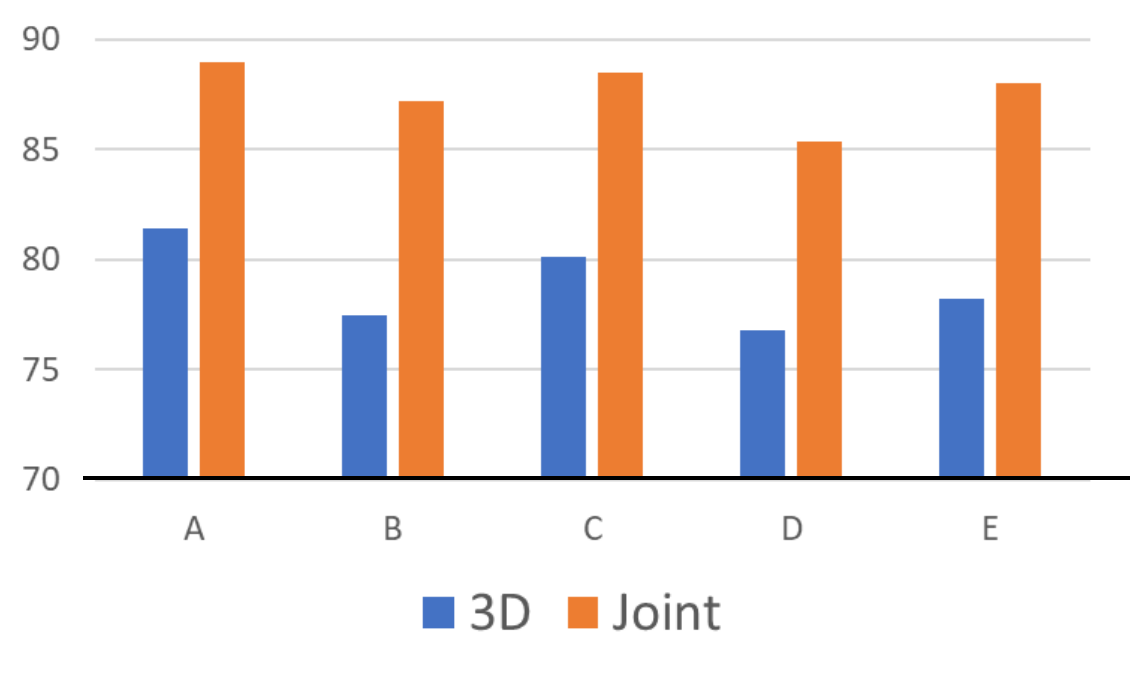}}
   \end{minipage}
   \caption {Performances with different augmentation strategy on 16-shot ModelNet40. Accuracy of 3D branch (\textit{Blue}) and \textsc{InvJoint} (\textit{Orange}) are reported. }
   \label{fig:d}
\end{figure}

As illustrated in~\cite{goyal2021revisiting}, auxiliary factors like different evaluation schemes, data augmentation strategies, and loss functions, which are independent of the model architecture, make a large difference in point cloud recognition performance. In order to test the robustness of \textsc{InvJoint}, we conduct different type of augmentations strategies with DGCNN~\cite{wang2019dynamic} as our 3D branch encoder, and report the results of individual 3D branch as well as the joint prediction of \textsc{InvJoint}. In detail, Table~\ref{tp3} summarize the compared 5 types of augmentation strategies. We could find from Figure~\ref{fig:d} that though the choice of augmentation strategy greatly influences the performance of 3D branch in few-shot settings, the joint prediction maintains encouraging and stable enhancement thanks to the collaborative joint training.

% \subsection{Parameter Sensitivity Analysis} 
%  % The following sensitivity analyses were conducted in ModelNet40 with 16-shot settings. (1) We first observed the optimal $\lambda$ in Eq.5. The Top-1 Accuracy is \textit{87.32 / 87.90 / 88.94 / 86.23 \%} ($\lambda$ = 0.1 / 1 / 5 / 10). (2) The we Keep $\lambda$= 5,  the Top-1 Accuracy is \textit{ 85.47 / 88.94 / 88.65 / 84.10 \%} with an added weight ratios ($\alpha$ = 0.1 / 1 / 5 / 10) between $\mathcal{L}_{C E}$ and $\mathcal{L}_{\text{align}}$, which vividly indicates that the robustness of \textsc{InvJoint} with low parameter sensitivity.

\subsection{Parameter Sensitivity Analysis}
The following sensitivity analyses were conducted in ModelNet40 with 16-shot settings. (1) We observed the optimal $\lambda$ in Eq.5. The Top-1 Accuracy is \textit{87.32 / 87.90 / 88.94 / 86.23 \%} ($\lambda$ = 0.1 / 1 / 5 / 10). (2) Keeping $\lambda$= 5,  the Top-1 Accuracy is \textit{ 85.47 / 88.94 / 88.65 / 84.10 \%} with an added weight ratios ($\alpha$ = 0.1 / 1 / 5 / 10) between $\mathcal{L}_{C E}$ and $\mathcal{L}_{\text{align}}$.

\begin{table}[]
\caption{Different Augmentation Strategies.}
\vspace{0.1cm}
\centering
\resizebox{0.465\textwidth}{!}{%
\begin{tabular}{  c | c }
\toprule[1.4pt]
 ID &   Augmentation Strategy     \\ \midrule[1.2pt]
 A                 &  \textbf{Random translation} \& \textbf{Random Scaling}     \\
B      &  \textbf{Jittering} \& \textbf{Random Rotation Along Y-axis}      \\
C         &  A \& \textbf{RandomInputDropout}  \&  \textbf{Random Rotation}   \\
D                &  B \& \textbf{RotatePerturbation}     \\
E     &    A \& B  \& \textbf{RandomInputDropout}   \\

\bottomrule[1.4pt]
\end{tabular}}
\label{tp3}
\end{table}

\subsection{Additional Results on HEM methods in Step 1} The GMM~\footnote{GMM-based loss discrimination has been widely adopted to identify outliers in de-noising and hard example mining because of its efficiency and high compatibility.} module is only used for selecting hard samples in each modality and is \textit{\textbf{NOT} our technical contribution}. Therefore, we further replace it with other loss discrimination and 
 OHEM (online hard example mining) methods for ablation. Specifically in Table~\ref{tp2}, if we replace our GMM module with Beta Mixture Model (BMM) in ModelNet40 with different few-shot settings, \textsc{InvJoint} can still achieve comparative results and largely outperform other methods in different few-shot settings.

 \begin{table}[t]
\caption{Few-shot performance on ModelNet40 with different OHEM methods. }
\vspace{0.12cm}
\centering
\resizebox{0.48\textwidth}{!}{%
\begin{tabular}{c | c | c c c c c }
\toprule[1.4pt]
Method                     &  OHEM     &  1-shot  & 2-shot & 4-shot & 8-shot & 16-shot     \\ \midrule[1.2pt]
PointCLIP~\cite{zhang2022pointclip}      & - &  52.96   & 66.73  & 74.47  & 80.96  & 85.45 \\
   Crosspoint~\cite{afham2022crosspoint} & - & 48.24   & 59.95  & 64.25  & 75.75  & 79.70   \\

\textsc{InvJoint}                         &  GMM    &  {68.85}   & {70.24}  & \textbf{78.95} & \textbf{82.85} &   \textbf{88.94}    \\ 
\textsc{InvJoint}                          &  BMM    &  \textbf{70.12}   & \textbf{71.30}  & {76.08} & {82.63} &   {87.15} \\
\bottomrule[1.4pt]

\end{tabular}}
\label{tp2}
\end{table}

\begin{table}[]
\caption{The enhancement ability of 2D branch in \textsc{InVJoint}.}
\centering
\resizebox{0.46\textwidth}{!}{%
\begin{tabular}{c | c | c| c }
\toprule[1.4pt]
  Methods      &      Before Fuse  & After Fuse  & Gain    \\ \midrule[1.2pt]
  PointNet++~\cite{qi2017pointnet++}        &   80.2  &  85.1   &   \textcolor[rgb]{0,0,1}{4.9}  \\ 
   CrossPoint~\cite{afham2022crosspoint}   &  79.7   &   83.2   &   \textcolor[rgb]{0,0,1}{3.5}  \\
  DGCNN~\cite{wang2019dynamic}        & 81.4  &  89.0   &  \textcolor[rgb]{0,0,1}{7.6} \\ 
  CurveNet~\cite{muzahid2020curvenet}   &   81.8   &  87.3   & \textcolor[rgb]{0,0,1}{5.5}  \\ \bottomrule[1.4pt]
\end{tabular}}
\label{tpp4}
\end{table}

\subsection{Additional Results on 3D Backbones} 
To verify the complementarity and the coordination of 3D and 2D, we further aggregate the fine-tuned 16-shot 2D branch on ModelNet40 with different 16-shot 3D backbones, including PointNet++~\cite{qi2017pointnet++}, CrossPoint~\cite{afham2022crosspoint}, DGCNN~\cite{wang2019dynamic}, CurveNet~\cite{muzahid2020curvenet}. Table~\ref{tpp4} illustrates that the enhancement ability of 2D branch in \textsc{InvJoint} with the alleviation of modality conflict.

{\small
\bibliographystyle{ieee_fullname}
\bibliography{egbib}
}

\end{document}